\documentclass[10pt,man,floatsintext]{apa7}
\usepackage{soul}
\usepackage{amsmath, amssymb}
\usepackage{bm}
\usepackage{graphicx}
\usepackage{subcaption}
\usepackage{hyperref}
\usepackage{todonotes}
\usepackage{float}
\usepackage[justification=centering]{caption}
\usepackage{pslatex}
\usepackage{apacite}
\usepackage{apacdoc}
\usepackage{lmodern}
\usepackage{tgcursor}

\newtheorem{definition}{Definition}
\newtheorem{example}{Example}

\usepackage{listings}
\DeclareCaptionFormat{listing}{\rule{\dimexpr\textwidth+17pt\relax}{0.4pt}\par\vskip1pt#1#2#3}
\newcommand{\red}[1]{ #1}

\title{Shades of confusion: Lexical uncertainty modulates \textit{ad hoc} coordination in an interactive communication task}

\authorsnames[{1,2},{1,3}, {1}]{Sonia K. Murthy, Thomas L. Griffiths,  Robert D. Hawkins}
\authorsaffiliations{{Department of Psychology, Princeton University, Princeton, NJ}, {Allen Institute for Artificial Intelligence, Seattle, WA}, {Department of Computer Science, Princeton University, Princeton, NJ}}
\shorttitle{Shared color associations}
\authornote{Materials and code for reproducing all behavioral experiments and analyses are open and available online at \texttt{\href{https://github.com/hawkrobe/color-ref}{https://github.com/hawkrobe/color-ref}}.
The pre-registration for Experiment 1 is  at \texttt{\href{https://osf.io/v86xz}{https://osf.io/v86xz}} and the pre-registration for Experiment 2 is at \texttt{\href{https://osf.io/gmqsv}{https://osf.io/gmqsv}}. Correspondence should be addressed to Robert Hawkins (\texttt{rdhawkins@princeton.edu}).}

\abstract{
There is substantial variability in the expectations that communication partners bring into interactions, creating the potential for misunderstandings. To directly probe these gaps and our ability to overcome them, we propose a communication task based on color-concept associations. In Experiment 1, we establish several key properties of the mental representations of these expectations, or \emph{lexical priors}, based on recent probabilistic theories. Associations are more variable for abstract concepts, variability is represented as uncertainty within each individual, and uncertainty enables accurate predictions about whether others are likely to share the same association. In Experiment 2, we then examine the downstream consequences of these representations for communication. Accuracy is initially low when communicating about concepts with more variable associations, but rapidly increases as participants form ad hoc conventions. Together, our findings suggest that people cope with variability by maintaining well-calibrated uncertainty about their partner and appropriately adaptable representations of their own.
}

\begin{document}

\maketitle

From jargon-filled scientific communication \cite{anderson2010hidden,bullock2019jargon,martinez2021specialized} and medical consultations \cite{korsch1968gaps,castro2007babel,mccabe2018miscommunication} to the linguistic battlefields of political discourse \cite{wodak1989language,lakoff2006whose}, it can often feel as if we are speaking different languages. 
Speakers not only bring different perspectives, expertise, and background knowledge into a conversation, they may even have different expectations about what words mean \cite{mccloskey1978natural,labov1973boundaries,marti2019same}.
For example, cognitive psychologists, doctors, and lawyers all commonly use the word ``trial'' in professional settings but refer to different concepts (a stimulus presentation, a clinical study, and a court appearance, respectively).
Such lexical variability is a troubling and pervasive challenge, setting the stage for misunderstandings and other communication breakdowns.
How do we manage to understand each other when we cannot be sure that we're starting on the same page?

Modern theories have addressed this challenge by viewing communication not as a unitary act of transmission but as an ongoing collaborative \emph{process} where interlocutors must coordinate to reach mutual understanding over time \cite{reddy1979conduit,davidson1986nice, clark1996using,krauss1996social,van-arkel-etal-2020-simple}.
These theories acknowledge idiosyncrasy, misunderstanding, and variation in literal meaning across partners and communities as a fundamental and unavoidable aspect of language use \cite{wilson2007unitary,clark1998communal,elman2004alternative,schuster2020know}. 
Still, a central open question has concerned the underlying mental representations of signal meaning that support this collaborative process. 
One hypothesis raised by recent probabilistic theories of communication is that the ability to anticipate and flexibly overcome misunderstandings depends on each speaker's initial \emph{lexical uncertainty}, reflecting prior expectations about what messages may or may not mean to one's partner \cite{potts2015negotiating,bergen2016pragmatic,brochhagen2020signalling,Hawkins2021FromPT}.
This idea builds on the classical construct of a mental lexicon containing signal-meaning mappings, allowing words to be grounded in external referents. 
For example, a speaker would consult their lexicon to evaluate the extent to which a given word like ``dog'' applies to a given animal they've encountered in the world.
Lexical uncertainty replaces a fixed dictionary of mappings with a probability distribution over \emph{possible} mappings that different partners may be using.
For example, there may be more uncertainty about whether a given partner will share a given meaning for some technical jargon than whether they will share a given meaning for ``dog\footnote{While lexical uncertainty is typically defined as a joint distribution over the full set of signal-meaning tuples $P(m, s)$ (see Appendix A), it can also be expressed as the conditional uncertainty over meanings for a given signal, $P(m | s)$, or as the conditional uncertainty over signals for a given meaning $P(s | m)$, since the baseline probabilities $P(m)$ and $P(s)$ are of lesser interest. We will loosely use \emph{lexical prior} and \emph{lexical uncertainty} to talk about this whole family of expressions, but the final conditional expression $(s | m)$ is particularly useful for eliciting priors.}.''

While lexical priors have played a key explanatory role in computational models of communication, they have been challenging to measure and manipulate directly in experimental work.
Classical studies of coordination and communication have typically been restricted to stimuli like ambiguous tangram shapes \cite{clark1986referring}, line drawings \cite{krauss1964changes}, or complex scenes \cite{weber2003cultural} where lexical priors fall in a narrow, carefully-calibrated band of uncertainty between completely random (e.g., white noise) and completely universal (e.g., a photograph of a prototypical dog).
Despite the narrow range of these stimulus spaces, it has been possible to isolate certain effects that are consistent with accounts of lexical uncertainty. 
For example, there is evidence that the \emph{codeability} of tangrams \cite<the number of distinct descriptions elicited;>[]{hupet1991effects}, and the shared expertise of speakers \cite<whether participants are equally familiar with New York landmarks;>[]{isaacs1987references} both affect the time it takes for speakers to reach mutual understanding. 
Still, there are many reasons to explore lexical uncertainty in richer stimulus spaces and other communication modalities.
Richer stimulus spaces make it possible to observe and manipulate lexical priors spanning a broader range of uncertainty, from the most idiosyncratic to the most universal. 
Other communication modalities present a further opportunity to overcome inherent challenges associated with natural language, where probability distributions must be estimated from sparse samples and poor coverage over the full (infinite) spaces of possible utterances and meanings. 
For example, measures of codeability for referential meaning \cite{hupet1991effects} are based on a single description from each participant, where most descriptions appear only once in the data set. 

\subsection{Color-concept associations as a window onto lexical uncertainty in communication}


\begin{figure}[h!]
\centering
\includegraphics[width=\linewidth]{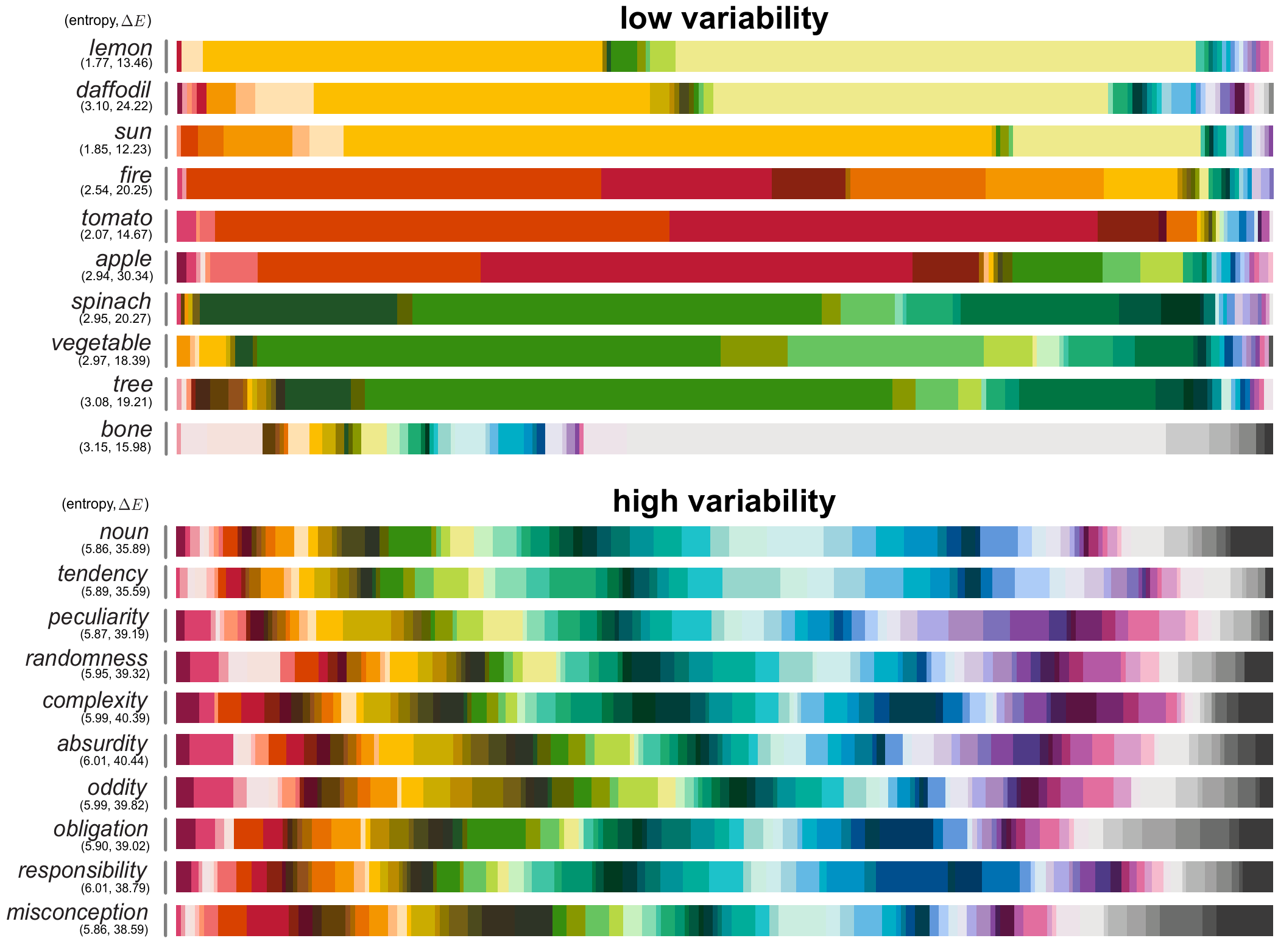}
\caption{\emph{Example color associations}. We depict the response distribution of the 10 concepts with lowest and highest variability. The width of each color bar corresponds to the proportion of a particular color response for a given word. Colors are presented in a fixed order across bars but colors that received no responses are not visible in the given bar. We report two estimates of variability for each concept (see Methods; in both cases, higher is more variable).}
\label{fig:colorresponses}
\end{figure}

To proceed in the face of these methodological challenges, we propose a communication paradigm based on color-concept associations.
While color has commonly been used as the \emph{target} of reference in communication tasks \cite{monroe2017colors,winters2019context,morin_muller_morisseau_winters_2020,caldwell2012cultural}, we instead ask participants to use a set of color chips as their \emph{communication modality} \cite<e.g.,>{roberts2018emergence} to communicate the identity of a target concept in a context of distractors (e.g. \emph{lemon}, \emph{happiness}).
While these color chips clearly differ in important ways from natural language utterances as a communication modality (see General Discussion), we argue that color-concept associations nevertheless provide a number of advantages for examining the consequences of lexical uncertainty.
First, color-concept associations naturally span a wide range of possible priors. 
For example, some concepts, like \emph{lemon}, are expected to have strong, nearly universal color associations, reflected in tight priors $P(\textrm{color} | \textrm{\textit{lemon}})$. 
Meanwhile, other concepts, like \emph{fairness} are expected to have more idiosyncratic and distributed associations, reflected in looser and more spread-out priors \cite{hutchings2004colour}.
Prior studies eliciting such associations have found systematic differences across different abstract concepts  \cite{rathore2019estimating,volkova2012clex,mohammad-2011-even,barchard2017sadness,Guilbeault2020}, which may also vary cross-culturally \cite{hupka1997colors,Tham2019ASI}.
We build on the elicitation methods developed in prior work while also introducing several methodological innovations to answer novel questions arising in the context of communication. 
Second, while natural language utterances are typically understood to be embedded in a complex and high-dimensional semantic space \cite{jones2007representing,pennington2014glove}, color signals are embedded in a much lower-dimensional space with better-validated psychometric structure, which allows for denser sampling and explicit measurements of variation. 
Finally, color remains an important modality of natural communication in its own right \cite{riley1995color}, as evidenced by the deliberate choice of color palettes in graphic design \cite{marcus1982color} and marketing \cite{marketingcolor}, or our everyday metaphorical appeals to color when trying to convey complex emotional states that are challenging to describe with words \cite{meier2005metaphorical,lakoff2008metaphors,van2011language}. 

\subsection{Three foundational questions about lexical uncertainty}

\begin{figure}[b]
\centering
\includegraphics[width=0.85\linewidth]{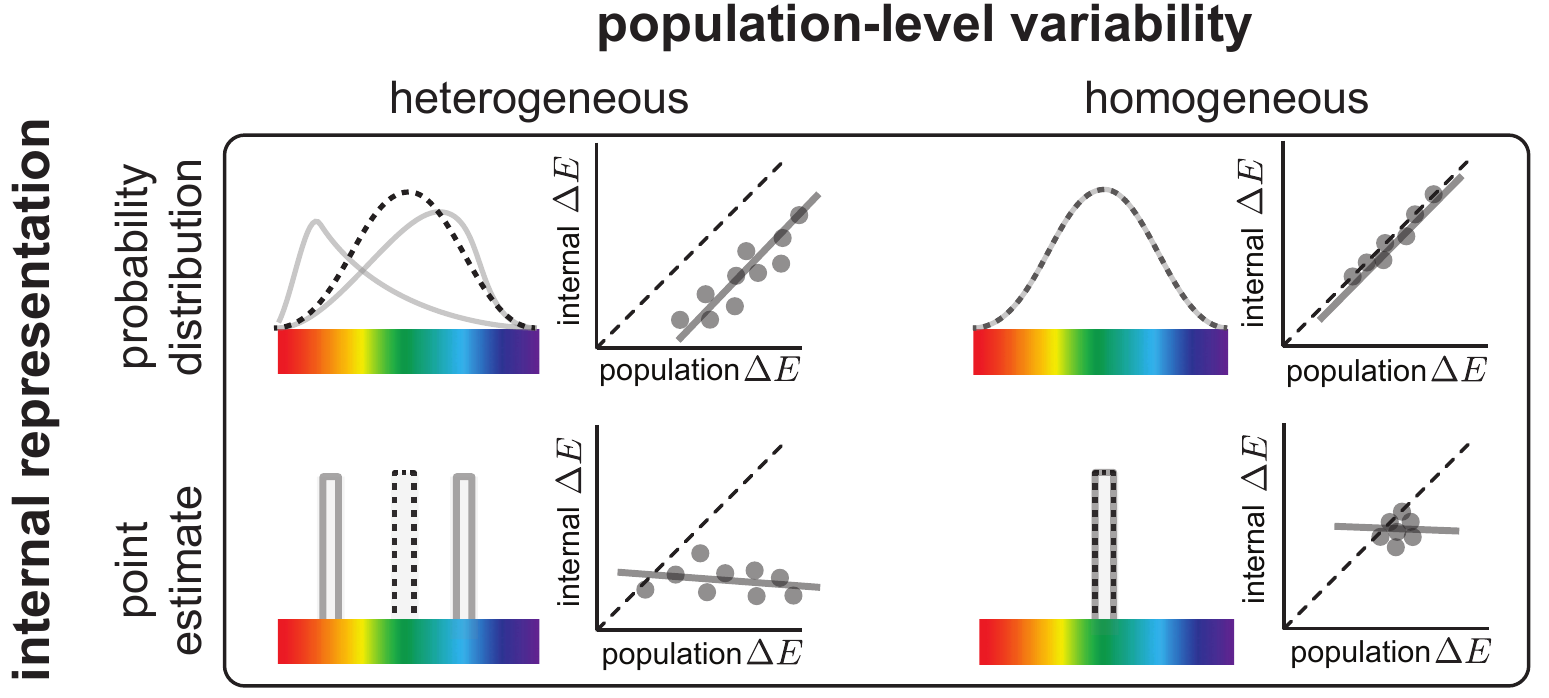}
\caption{\emph{Schematic of candidate hypotheses and corresponding predictions.} We explore a $2 \times 2$ space of proposals about how color-concept associations are internally represented by individuals (as a fully calibrated probability distribution representing uncertainty, or as a sparser point estimate) and how much these associations differ across individuals in the population (from a fully homogeneous population to a more heterogeneous population). These hypotheses make distinct predictions about the relationship between the variability of associations measured within an individual and across a population, which we evaluate in Exp.~1.}
\label{fig:schematic}
\end{figure}

We use the domain of color-concept associations to evaluate three foundational hypotheses about lexical uncertainty raised by recent theories of communication.
First, and most fundamentally, do individuals actually maintain an internal probability distribution representing their uncertainty about associations (Fig.~\ref{fig:schematic}, top row), or do they only represent a point estimate giving their strongest association (Fig.~\ref{fig:schematic}, bottom row)? 
Second, we ask: is the population relatively homogeneous, composed of individuals sharing similar representations (Fig.~\ref{fig:schematic}, right column) or is the population actually more heterogeneous and idiosyncratic  (Fig.~\ref{fig:schematic}, left column)?
Third, when it comes time to use these representations in a communicative context, is a given individual's representation purely \emph{egocentric} or do they maintain well-calibrated expectations about whether their representation will be shared by \emph{other} agents? 

While we unpack these hypotheses and operationalize their predictions more thoroughly in subsequent sections, it is worth noting here that there is theoretical precedent for these questions not only in the communication literature but also in the literature on color-concept associations.
For example, the \emph{color inference} framework introduced by \citeA{schloss2018color_inference} proposes that individuals store and continually update their color-concept associations from their experiences in the world.
Under this framework, every concept has a corresponding association space with some weight placed on each possible color.
These weights effectively give rise to an internal probability distribution that can, for example, be used to generate appropriately discriminative colors to convey different meanings \cite{mukherjee2021context}.   
Such resonances between models of meaning across the domains of natural language communication and of color-concept associations provides a further theoretical motivation for using the color domain to explore representations of lexical uncertainty.

In Experiment 1, we begin to explore basic theoretical properties of lexical uncertainty by eliciting color associations for a variety of concepts. 
By eliciting multiple responses from each participant, we were able to compare the \emph{population-level} variability of responses to the \emph{internal} variability within each individual.
And by asking participants to estimate the extent to which others will share the same associations, we were able to probe the extent to which any given individual represents the population distribution, thus forming a basis for successful communication. 
Next, in Experiment 2, we directly measure the downstream effects of lexical priors on \emph{ad hoc} coordination using an interactive communication task. 
Pairs of participants were asked to communicate about sets of concepts using color chips as signals. 
Critically, we used the corpus of associations we elicited in our first experiment to construct contexts that \emph{manipulated} participants' initial priors: some associations were strongly shared across the population while others were more idiosyncratic.
Taken together, our work suggests that people maintain well-calibrated prior expectations about the potential for miscommunication and use these flexible priors to rapidly adapt to their partner.

\section{Experiment 1: Eliciting color-concept association priors}

We begin by eliciting color association priors for concepts spanning a wide range of population variability.
While there are a number of existing datasets that may be used to assess variability at the population level \cite<e.g.>{mohammad-2011-even,volkova2012clex,Tham2019ASI}, there were three specific desiderata for our work that were not satisfied by these existing datasets.
First, we required participants to not only choose the color they themselves most associated with a concept, but also to predict whether that association would be shared by others. 
Such an explicit query about expected agreement was not included in the protocol of previous studies.
Second, while previous studies revealed differences in population-level variability, it remains unclear whether such variability arises within individuals (e.g. from internal probability distributions) or only at the population-level (e.g. from individuals with different point estimates). 
To distinguish between these possibilities, we required multiple blocks of judgements for each participant; previous variants of the task only collected a single best exemplar per participant. 
Finally, we required a set of stimuli that would span a wide range of expected association variability from strong (e.g. ``lemon'') to weak (e.g. ``fairness'').
Previous studies have focused on smaller sets of concepts \cite<e.g.>[used only 59 abstract concepts, screening out words that corresponded to concrete objects like ``lemon'']{Tham2019ASI} or only collected coarse-grained responses for basic-level colors (e.g. \citeNP{mohammad-2011-even} asked participants to choose from the 11 color terms from \citeNP{berlin1991basic}).

\subsection{Methods}

\paragraph{Participants}

We recruited 733 participants from Amazon Mechanical Turk, restricting location to the United States. 
After implementing our pre-registered exclusions criteria, we were left with data from 485 of these participants.~
129 participants were excluded for failing one of four attention checks that were interspersed throughout the experiment. 
These attention checks asked participants to provide color associations for basic-level color words (e.g., ``red'', ``orange'', ``yellow'').
The first two attention checks were embedded in a block of initial practice trials while the other two were inserted randomly into the experiment (one in the first half, and one in the second half).
Participants who did not provide a color within a (relatively permissive) set of valid responses were immediately removed from the experiment for the base payment. 
We also removed 5 additional participants who provided the same response for more than three trials in a row, or consistently responded in less than 1000ms, both of which indicated blind guessing without reading the stimulus prompt.
Finally, an additional 65 participants were excluded due to colorblindness. 
To test color vision, we presented participants with three Ishihara plates that detected common red-green deficiencies or more extreme colorblindness. 
We excluded only those participants whose responses indicated more extreme colorblindness (or inattention). 

\paragraph{Stimuli}

We considered two factors when selecting our stimulus set. 
First, we required a relatively large number of concepts to control for possible item effects.
Second, we needed these concepts to span a wide range of different priors (i.e. different color associations, with different levels of variability), but were unable to know these priors beforehand. 
We thus considered a set of 5,500 candidate concepts drawn from the Glasgow Norms dataset \cite{glasgow}, which provides ratings for words on 9 different scales, including properties like imageability and concreteness. 
Concreteness represents the degree to which something can be experienced by the senses, while imageability represents the extent to which a word invokes a mental image\footnote{Though these two measures are highly correlated with with each other ($r=0.93$), they are considered distinct aspects of a word's semantics \cite{paivio1968concreteness,richardson1976}. For example, emotion words like \emph{anger} may be rated low on concreteness but high on imageability; conversely, some scientific or medical concepts like \emph{diabetes} may be high on concreteness but low on imageability}.
We used these measures as rough proxies for the level of variability in color associations we could expect for a concept and selected a balanced set of 200 concepts from this candidate set using the following procedure.
We began by imposing a familiarity threshold (familiarity $\geq4.0$) to ensure that the majority of participants were likely to know the concept that each stimulus word referred to. 
Next, we selected the 500 words with the highest concreteness ratings (the \emph{concrete} set) and the 500 words with the lowest concreteness ratings (the \emph{abstract} set).
We then sampled 100 words from each set to obtain a roughly uniform distribution of imageability. 
Lastly, we conducted a manual pass over the resulting 200 words to ensure that they were consistent in part of speech (e.g., converting adjectives to their noun form) and to replace any that were offensive, confusing, or redundant with one another. 
These 200 words were randomly divided into 5 subsets, each containing 20 words from the \emph{abstract} set and 20 words from the \emph{concrete} set. 

\paragraph{Task, design, \& procedure}

\begin{figure}[t]
\centering
\includegraphics[width=\textwidth]{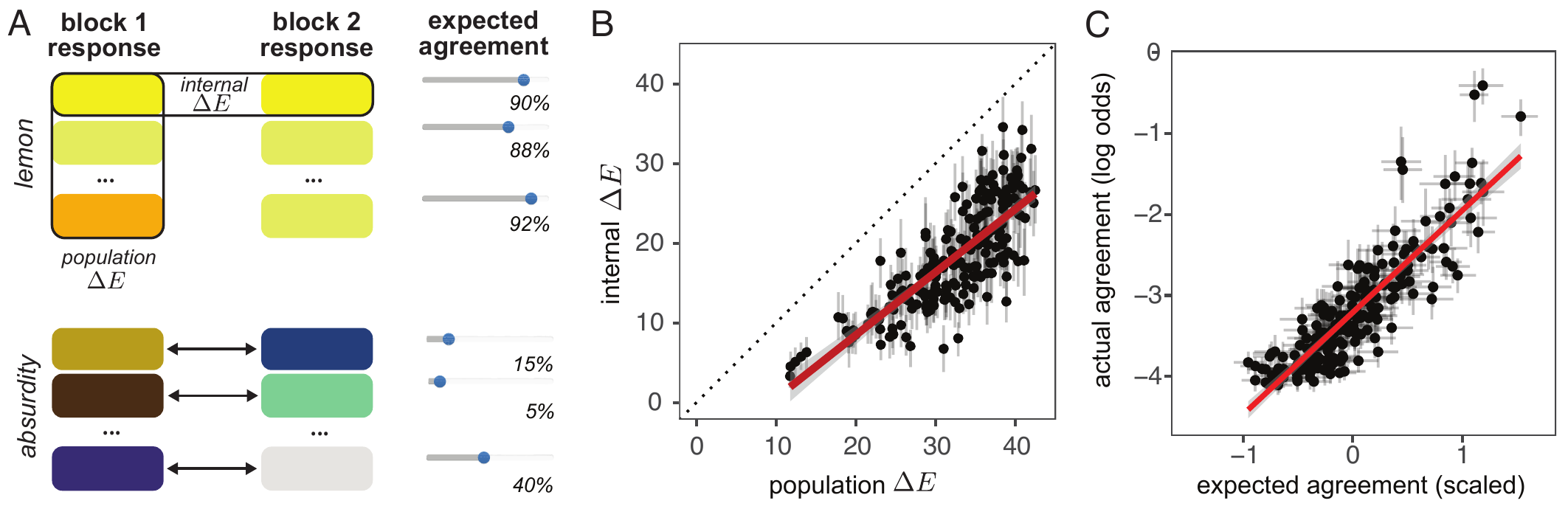}
\caption{\emph{Design and results of Experiment 1.} (A) We elicit two associations from each participant for a given word (1st block and 2nd block). On the second block, participants additionally estimate the probability that others would provide the same association. Internal $\Delta E$ is measured as the similarity between the two responses provided by the same individual, while population $\Delta E$ is measured between different individuals. (B) Internal $\Delta E$ of color associations for each concept tracks population variability $\Delta E$, although a single participants' responses are somewhat closer together than expected from sampling from the population distribution. (C) Expectations are well-calibrated to the true agreement odds. Error bars for each concept are bootstrapped 95\% CIs.}
\label{fig:exp1_results}
\end{figure}

Each participant was assigned one of the 5 distinct word sets.
On each trial, a single word was presented with a set of 88 (virtual) Munsell chips sampling a wide range of color space\footnote{The World Color Survey \cite{kay2009world} used the set of 320 chips (40 evenly spaced hues crossed with 8 levels of lightness) proposed by \citeA{lennebergroberts}. Later studies down-sampled this set to 160 \cite{heider1972universals} by removing hues at intermediate levels of 2.5 and 7.5, and then further down-sampled to 80 \cite{gibson2017color,zaslavsky2019color}. Our 88-chip set was derived by adding 8 achromatic chips to the 80-chip set from \citeA{gibson2017color}, allowing participants to select greyscale values.}.
Participants were instructed to click the color they most associated with the target word. 
To control for differences in individuals' color displays, participants were instructed to take the experiment on a desktop or laptop computer and ensure that their screens were set to their default brightness and color temperature (e.g. to turn off programs like Flux). 
Participants were also screened through a pre-test asking them to select color swatches for words like ‘blue’ and ‘red’, ensuring that any differences in color displays lay within tolerance of color boundaries (see Supplemental Figs. \ref{fig:colorblind-trials}-\ref{fig:instructions-experiment1}).
To estimate internal variability, we presented these words in a blocked sequence. 
After providing responses for all 40 words in the set, presented in randomized order, participants repeated the task a second time (Figure  \ref{fig:exp1_results}A). 
On the second block, participants were also asked: ``How strongly do you expect others to share your color association for this word?'' 
We presented a slider ranging from ``not at all'' (most people will have a different color association than I do), to ``very strongly'' (most people will have the same color association as I do) with a midpoint labeled ``somewhat'' (roughly the same number of people will have the same or different color associations as I do).
This question allowed us to compare the true proportion of shared responses to each participants' expectations.

\paragraph{Evaluation Metrics}
We measured variability in two different ways: (1) the entropy of the discrete response distribution and (2) the average pairwise similarity between responses in color space\footnote{We also pre-registered a procedure to estimate parametric variability by fitting a multi-dimensional Gaussian distribution over color space, but chose to replace this measure by the $\Delta E$ measure. It is highly correlated with the Gaussian measure, $r = 0.861$, but better accounts for metric distortions in color space and the existence of multi-modal response distributions.}. 
Our entropy measure was computed on the distribution of response counts over the 88 Munsell chips, using the Schurmann-Grassberger estimator (i.e. adding pseudo-counts of 1/88 to each bin). 
Entropy is expected to be high when responses are spread out across many different color chips, and low when participants all concentrate their responses on a small number of colors. 
Our pairwise distance measure is computed by taking the perceptual similarity $\Delta E$ (where $E$ stands for \emph{Empfindung}, German for ``sensation'') between colors in the psychometrically-validated CIELAB color space.
It is close to 0 when colors are perceptually similar (colors with $\Delta E < 1$ are not able to be discriminated by the human eye), and reaches values close to 100 for extremely dissimilar colors.
We use the CIE2000 definition of $\Delta E$, which accounts for distortions in perceptual uniformity \cite{cie2000analysis}.
For each word, we derived a measure of internal variability, \emph{internal} $\Delta E$, by taking the $\Delta E$ between the color chosen by a given participant in the first block and the second block.
We also obtained a word-level measure of population variability, \emph{population} $\Delta E$, by taking the average pairwise $\Delta E$ among every pair of color responses provided by different participants\footnote{We may define population $\Delta E$ in several ways, depending on where the different participants' responses are drawn from. For example, block 1 $\Delta E$ compares responses from different participants in the first block $\Delta(c_{1i}, c_{1j})$, block 2 $\Delta E$ restricts to the second block, $\Delta(c_{2i}, c_{2j})$, and a ``cross-block'' $\Delta E$ compares responses in one block to those obtained from other participants in the other block, $\Delta(c_{1i}, c_{2j})$. 
These different ways of measuring population $\Delta E$ are highly correlated ($r=0.994$ between the block 1 and cross-block versions; $r=0.993$ between the block 2 and cross-block versions; $r=0.973$ between the block 1 and block 2 versions), and results are invariant to this choice. While Fig.~\ref{fig:exp1_results}A shows the block 1 variant to make it clear that the axis of comparison is across different participants, we report results using the version that pools together all responses across both blocks to get the most highly powered estimate.}.

\subsection{Results}

\paragraph{Associations are more variable for abstract words}

We begin by examining our use of concreteness and imageability as proxies for constructing stimuli that elicit a wide range of associations. 
These proxies were motivated by recent work suggesting that the extent to which people share associations for a concept may be related to the extent that they share the same \emph{sensory experiences} for that concept, such that abstract concepts have more variable associations than concrete and imageable concepts.
For example, recent analyses of Google Image search results have found that raw color distributions found in the top images retrieved for abstract terms are indeed more variable on average than those measured for concrete terms \cite{desikan2020comp}.
At the same time, however, the relationship between these measures is likely more complex: recent work has also found coherent relationships among abstract concepts \cite{Guilbeault2020} and some abstract words may nevertheless have strong associations (e.g. \emph{anger} is red and \emph{sadness} is blue).
To assess this relationship in our dataset, we constructed two regression models predicting our two population variability measures (entropy and $\Delta E$) from the corresponding concreteness and imageability ratings reported by \cite{glasgow} for each word (see Appendix B, Fig.~\ref{fig:relationships}).

We found that both concreteness and imageability are independently correlated with entropy (Spearman rank correlation $\rho = -0.51$, $p<0.001$ and $\rho = -0.54$, $p < 0.001$, respectively), suggesting that color associations for more abstract words were much more variable.
However, in a combined model including both predictors we only found a main effect of imageability ($b = -0.31$, $t(191) = -3.19$, $p = 0.002$), with no independent contribution of concreteness ($b= 0.03$, $p=0.74$), suggesting that the information provided by concreteness may be redundant.
Similarly, both concreteness and imageability are independently correlated with $\Delta E$ ($\rho=-0.53$, $p<0.001$ and $\rho=-0.52$, $p<0.001$, respectively), but we found the opposite relationship in the combined model: concreteness was a weak but significant predictor of $\Delta E$ ($b = -1.22$, $t(191)= -2.07$, $p = 0.039$), while imageability was not ($b= -0.8,~p=0.26$). 
Together, these findings support a strong negative relationship between abstractness and variability, regardless of how we measure them, but we were unable to distinguish the unique contributions of concreteness from those of imageability due to their high collinearity in the combined model.
Importantly, while they were useful for choosing stimuli, neither measure was a particularly good proxy for population variability in absolute terms.
We therefore use our own direct estimates of variability (i.e. entropy and $\Delta E$) to construct high and low variability conditions in Experiment 2, rather than relying on coarser concreteness or imageability norms.

\paragraph{Population variability reflects individual uncertainty}

Given that we successfully elicited associations with different levels of variability in the population, we now proceed to ask how this variability is related to the internal representation of color-concept associations within each individual.
First, we ask whether individuals maintain a single \emph{point estimate} representing a single strongest association or instead maintain a full \emph{internal probability distribution} over different associations.
Second, we ask whether the population is relatively \emph{homogeneous}, where all individuals maintain roughly the same representations, or whether it is more \emph{heterogeneous}, containing individuals with somewhat differing representations (see Fig.~\ref{fig:schematic}).
To evaluate this space of candidate hypotheses, we leveraged our blocked design to compare \emph{internal} variability against \emph{population} variability.

In particular, we draw on the notion of the ``crowd within'' \cite{vul2008measuring,rauhut2011wisdom,hourihan2010smaller,steegen2014measuring,van2018wisdom,fiechter2021wisdom} that has recently been proposed for phenomena like forecasting and estimation in the decision-making literature \cite<see>[for a review]{herzog2014harnessing}.
The ``crowd within'' views a judgement as a sample from an (implicit) probability distribution, explaining why the same participant may make different judgements at different times, and why averaging together multiple judgements from the \emph{same} participant may yield better predictions (an analogue of the ``wisdom of the crowd'' where judgements are averaged across different participants). 
Thus, while it is unrealistic to perfectly reconstruct any single individual's prior distribution from only two responses per concept, a key insight of this literature is that we do not need to reconstruct the \emph{full} distribution maintained by any single participant. 
It is statistically sufficient to compare the distance between a small number of samples obtained from a given participant for a given concept (internal $\Delta E$) against the expected null distribution of distances derived from different participants for that concept (population $\Delta E$; see Figure \ref{fig:exp1_results}A).

Each hypothesis predicts a different pattern of relationships between internal $\Delta E$ and population $\Delta E$ (see Fig.~\ref{fig:schematic}). 
The distinction between different internal representations concerns whether internal and population variability are related across concepts. 
The \emph{point estimates} representation allows for some response noise but assumes such noise is identical across all concepts, hence internal variability should be independent of population variability and we would expect similar internal $\Delta E$ across blocks for all concepts\footnote{More formally, under this hypothesis, we could suppose an individual $i$'s responses $j$ are expected to be drawn from $r_{ij} \sim \mathcal{N}(c_i, \epsilon)$ where $c_i$ is their point estimate for the concept and $\epsilon$ captures a fixed probability that they may click on nearby color chips rather than the precise value $c_i$.}.
Meanwhile, the \emph{internal probability distribution} account predicts that individuals maintain different distributions for different concepts, which vary in their internal variability, hence there should be a systematic relationship between internal $\Delta E$ and population $\Delta E$ (regardless of whether the population is more homogeneous or heterogeneous).
The distinction between population homogeneity and heterogeneity, on the other hand, concerns the extent to which population variability is higher or lower overall than internal variability. 
In the extreme case where every participant $i$ shared the same distribution of color associations $P(r | c)$ in a perfectly homogeneous population, then the entire dataset of responses $r_{ij}$  would be drawn from the same distribution $r_{ij} \sim P(r | c)$ and the distances between two samples taken from the same participant $i$, $\Delta(r_{i1},r_{i2})$, would be the same, in expectation, as the distances between samples taken from distinct participants, $\Delta(r_{1j},r_{2j})$. 
In other words, we may obtain a distribution of the expected variation in any pair of sampled responses, under the null hypothesis of a homogeneous population, and compare the extent to which the actual pairs of samples we obtained from the same participant are more or less similar than expected under this null\footnote{There exist other possible explanations for a smaller internal $\Delta$E, although we have taken measures to minimize them; for example, if participants’ response was strongly influenced by their response on the first block, then their samples may violate the assumption of independent draws from the distribution. However, such influence is unlikely given the number of intervening stimuli in each block and the relatively long delay between samples. We return to this concern in the discussion.}.

Our findings are shown in Fig.~\ref{fig:exp1_results}B.
We found that the average $\Delta E$ between an individual participant’s responses was strongly correlated with the overall population's $\Delta E$, $r = 0.81,~t(192) = 19.04,~p < 0.001$. 
That is, for concepts where the population as a whole most disagreed with one another, each individuals' own responses across blocks also tended to disagree with one another. 
This pattern was only predicted by the \emph{internal distribution} account and not the \emph{point estimates} account, which cannot explain why there would be such systematic differences in internal $\Delta E$ across concepts. 
At the same time, we found that internal variability was lower than population variability overall; the estimated intercept in a linear regression significantly differed from zero, $b=-17.3$, $t(192)=20.8$, $p <0.001$, consistent with a more heterogeneous account where not all participants shared exactly the same internal distribution. 
A homogeneous population would be closer to the line of unity. 

\paragraph{Expectations about others are well-calibrated}

Our results so far support the view that each individual implicitly maintains a full probability distribution or \emph{prior} for each concept, which tracks the true population variability but which tends to be somewhat narrower on average. 
However, it is unclear whether this representation has a social component. 
In principle, communicative success depends on each individual's expectations about how their \emph{partner} will understand (or misunderstand) a given message, not just whether individuals represent similar meanings. 
This distinction is crucial for understanding the consequences of lexical priors for resolving misunderstandings in communication. 
If individuals are unable to represent whether \emph{others} will share the same prior (whether it is a point estimate or a distribution), they might over- or under-estimate agreement and inhibit successful communication\footnote{Of course, there are multiple ways for well-calibrated social expectations to be achieved. 
One possibility is that individuals maintain their own idiosyncratic associations, based on their unique experiences, but also track the degree of divergence from the overall population and correct for it in social settings using theory of mind (i.e. they are aware that their associations are idiosyncratic). 
Another possibility is that agents are egocentric (i.e. maintain a single internal distribution of associations) but have tuned their own internal distribution over time to match the population distribution.
Distinguishing between these representational possibilities is outside the scope of this paper.}.
To assess whether participants maintain well-calibrated social expectations, we turned to the expected agreement measure collected on the second block.
For each expected agreement rating in our dataset, we calculated the log odds that other participants' color responses actually matched that participants' reported color association, $\log(p / (1-p))$, representing ground-truth agreement.
We found a strong correlation between expected agreement and true agreement (Pearson's correlation $r = 0.88, t(197) = 25.5, p <0.001$ and Spearman's rank correlation, $\rho=0.87, p < 0.001$), suggesting that individuals' expectations about the others' were remarkably well-calibrated to the true statistics (see Figure \ref{fig:exp1_results}C)\footnote{While the log-odds linking function is orthogonal to our question of interest, we note that it is consistent with previous observations about how participants spontaneously map sliders to logarithmic rather than linear scales \cite<e.g.>{griffiths2005structure,griffiths2007mere,landy2013estimating}.}.

\section{Experiment 2: The consequences of lexical priors for communication}

Experiment 1 established several key properties of word-color associations: more abstract words have more variable associations, this variability is represented within individuals, and individuals can accurately predict whether an association is likely to be shared with others. 
These properties are precisely those that probabilistic models of communication have associated with \emph{lexical uncertainty}, the recognition that particular word meanings or concepts may or may not be shared with others.
Such uncertainty may allow for more flexible adaptation than simple point estimates: individuals may be able to anticipate potential confusions ahead of time, and possess rich enough knowledge about likely alternative associations to rapidly adjust their expectations on the fly.
Here, we examine the downstream effects of these priors in a Pictionary-like communication task where participants sent color swatches as messages that allowed their partner to guess a target word. 
We hypothesized that target words with nearly universal color associations, reflected in strong, tightly overlapping priors, would provide a common foundation and allow for instant communicative success. 
Meanwhile, words with more variable or uncertain color associations would be more difficult to communicate about. 
In either case, individual pairs should still be able to adaptively coordinate on mutually agreeable solutions given the flexibility provided by their initial uncertainty. 

\subsection{Methods}
\paragraph{Participants}
We recruited 234 participants on Amazon Mechanical Turk and paired them up to form 117 dyads. 
After removing 6 dyads that disconnected before completing the task and 4 additional dyads where at least one participant failed our attention checks, we were left with 107 dyads in our sample.
Participants were screened for comprehension and color vision before being paired with a partner (see Supplemental Fig.~\ref{fig:exp2-comprehension}). 

\begin{figure}[t]
\centering
\includegraphics[width=\textwidth]{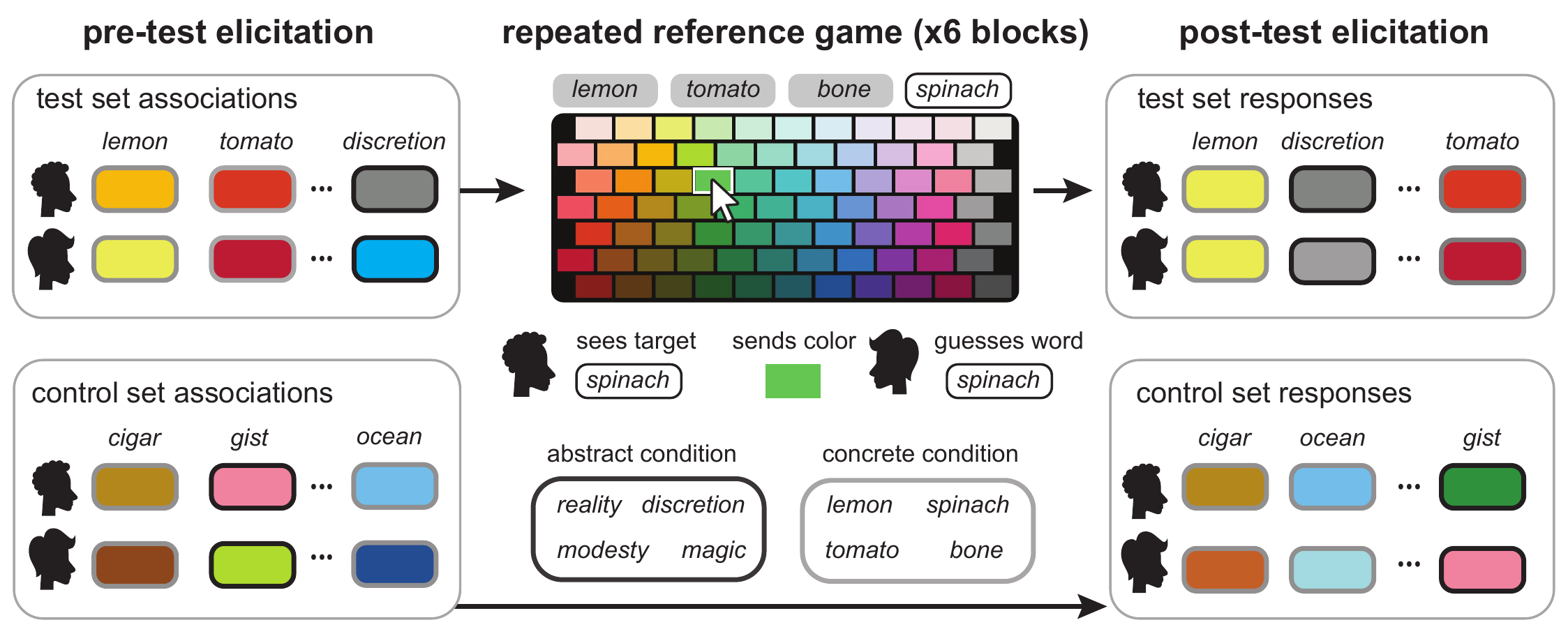}
\caption{\emph{Design of Experiment 2.} Participants are paired into dyads and assigned to \emph{speaker} and \emph{listener} roles. \red{Both participants are shown the full set of Munsell chips and the same context of 4 words}. The speaker is \red{additionally} shown \red{which one of these 4 words is} the target word and asked to select a color to send to the listener. The listener then guesses the word they believe is the target. Participants independently complete a color association elicitation task before and after the reference game for the \emph{test set} of eight words appearing in the communication game and also a \emph{control set} of eight additional words.}
\label{fig:exp2_design}
\end{figure}

\paragraph{Stimuli}
We constructed 100 four-word contexts from the 200 words we used in Experiment 1. 
We designed these contexts to span a broad range of priors based on the population variability we estimated in the prior experiment.
We iteratively sampled contexts to satisfy to two main constraints, both derived from possible pragmatic context effects which may mask the effects of priors: (1) we aimed to group words with similar variability while (2) minimizing the extent to which different words in the same context have overlapping priors.
To satisfy these constraints, we first ordered words by their estimated response entropy and greedily sampled from the list of words to build an initial context.
To check the extent to which these words had overlapping priors, we computed the Jensen-Shannon (JS) divergence between the Experiment 1 response distributions for each pair of words in the proposed context. 
We imposed a minimum divergence threshold of 0.3: when the context exceeded this threshold for a word, we replaced it with the next in the list until a satisfactory set was formed. 
We repeated this procedure to obtain 50 contexts where each word appeared in exactly one context and overlap within each context was low.
To obtain a distinct alternative set of contexts, we repeated the same procedure with the additional criterion of rejecting pairs of words that had previously appeared together in the first set.
These 50 contexts were divided into a ``high prior variability'' condition (the top 25) and a ``low prior variability'' condition (the bottom 25).

\paragraph{Procedure}

Participants were paired into dyads to play an interactive reference game using color as the communication medium (Fig.~\ref{fig:exp2_design}). 
Participants were randomly assigned to \emph{speaker} and \emph{listener} roles and placed in an environment containing a context of 4 concept words and 88 Munsell chips, both shared in common ground. 
At the beginning of each trial, one of the 4 words was privately shown to the speaker as the target word.
The speaker was then instructed to choose a color chip from the set of Munsell chips that would best allow the listener to select that target from the distractors.
After the listener received the speaker's message and clicked on one of the words, both participants received feedback: the listener was shown the true target and the speaker was shown the listener's selection.
Participants were awarded a performance bonus of \$0.03 for each correct response.

\paragraph{Design}

We tested the effect of lexical priors by manipulating the target words in a within-dyad design: each dyad was assigned two 4-word contexts, one ``high prior variability'' context and one ``low prior variability'' context.
The trial sequence was constructed from 6 repetition blocks, allowing us to observe the trajectory of behavior as each target is referred to multiple times.
The four target words from each context were randomly interleaved in each block, for a total of 48 trials (six blocks of eight words).
Participants switched roles between repetition blocks.
Finally, we included two blocks of the prior elicitation task used in Experiment 1. 
At the beginning and end of the reference game, we asked participants to provide associations for the 8 \emph{test} words in the contexts they were assigned as well as 8 \emph{control} words not encountered during the reference game task -- four with high prior variability and with low prior variability.

\subsection{Results}

\begin{figure}[t!]
\centering
\includegraphics[width=.8\textwidth]{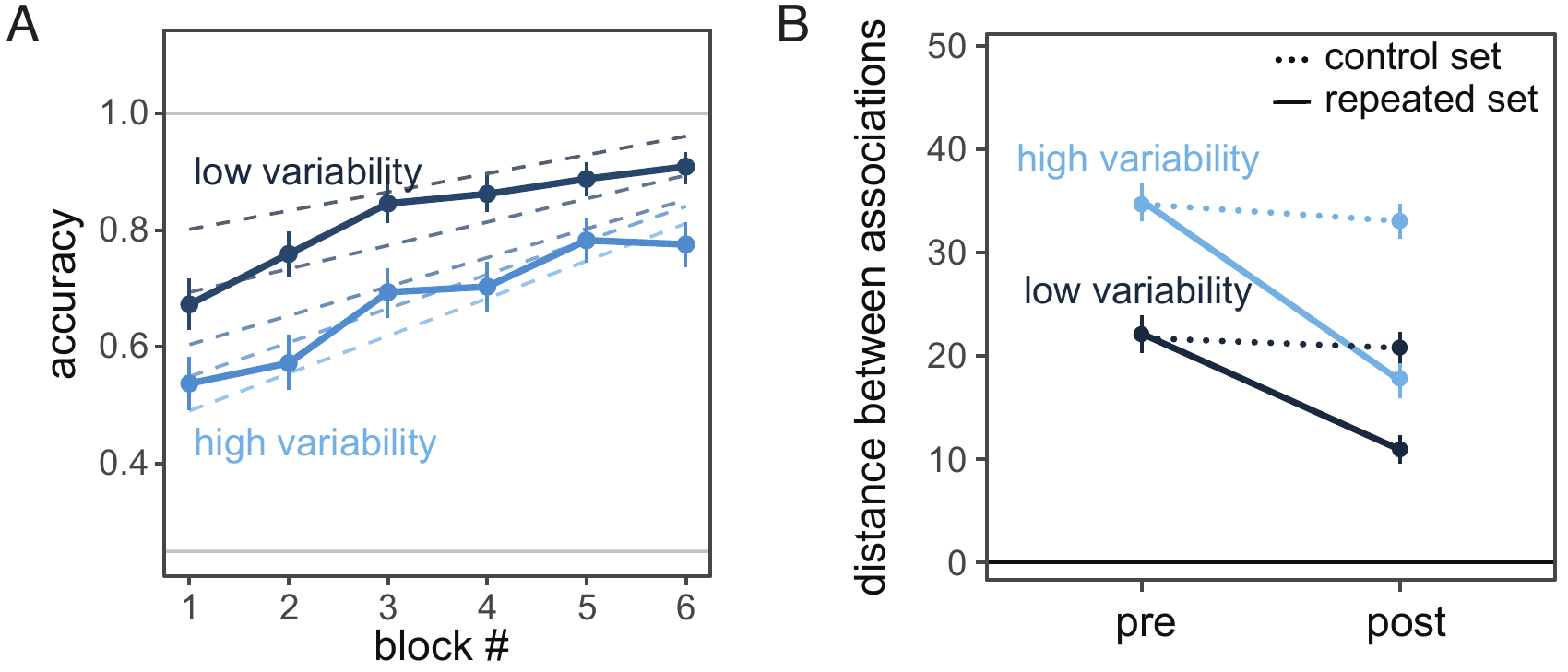}
\caption{\emph{Results of Experiment 2.} (A) Communicative success increases across repeated interaction. Dashed lines represent quintiles of initial similarity between partners' associations in the pre-test (a finer-grained measure than condition). (B) Speakers' associations begin closer in color space for low prior variability words than for high prior variability words, but converge to closer associations in the post-test, relative to a control set that did not appear in the reference game. Error bars are bootstrapped 95\% CIs.}
\label{fig:exp2_results}
\end{figure}

\paragraph{Shared priors facilitate communicative success}

Our first prediction concerned the effect of priors on communicative success, which we operationalized as the probability that the listener correctly selects the target (chance is 25\%).
We hypothesized that participants would initially struggle more to communicate in the high prior variability condition, compared to the low prior variability condition, where they could take advantage of stronger expectations and more closely overlapping priors.
At the same time, we expected participants to improve through interaction, as they built common ground across repeated appearances of the same targets.
To test these hypotheses, we constructed a logistic regression to predict correctness at the trial-by-trial level, including fixed effects of condition (high prior variability context vs. low prior variability context), repetition block number, and their interaction, with the maximal random-effect structure that converged (random intercepts and both main effects at the dyad-level).
We found a significant main effect of condition ($b=0.44,~z = -4.8,~p < 0.001$) with higher accuracy for the low prior variability condition throughout the task.
We also found a significant improvement in accuracy across the task in both conditions ($b=0.43,~z = 12.0,~p < 0.001$), reflecting \emph{ad hoc} coordination.
Finally, these effects were clarified by a significant interaction ($b=0.06,~z=-2.6,~p = 0.009$), likely reflecting ceiling effects for the low-variability conditions. 
On the first round, dyads were at approximately 54\% accuracy in the \red{high} variability condition, compared to 67\% accuracy in the \red{low} variability condition.
By the final round, they achieved approximately 78\% and 91\% accuracy, respectively\footnote{A Bayesian logistic regression model with the maximal random-effect structure at both the dyad- and item-level yielded similar effects (see Appendix Figure \ref{lst:bayesianmodel} for full results)}. 
Because each condition contained a wide variety of items spanning different priors (reflected in high variances for random effect  estimates), we also probed these effects using a more fine-grained, individualized measure. 
We computed the $\Delta$E distance between the two participants' pre-test responses for each word in each game, and found that quintiles of this continuous measure followed the same trend (dashed lines in Figure \ref{fig:exp2_results}A).

\paragraph{Rapid convergence to shared meanings}

To evaluate the extent to which participants flexibly shifted their associations for target concepts over the course of interaction, we compared pre-test and post-test responses. 
We operationalized the similarity between partners' associations as the $\Delta E$ between their responses at each phase. 
For example, as a manipulation check, we found that participants' responses were indeed more similar in the pre-test for words in the low variability condition than the high variability condition, $d = 12.78$, $t(117.3) = 9.1$, $p<0.001$), implying that we successfully constructed separable context sets from the population variability estimates obtained in Experiment 1.
Critically, however, we hypothesized that participants' responses in the post-test would become significantly closer for the words that were repeated in the reference game, compared to a comparable set of control words that only appeared in the pre-test and post-test.
We tested these predictions in a mixed-effects linear regression model including fixed effects of phase (pre-test vs. post-test), word set (repeated vs. control), and condition (high vs. low variability) as well as all interactions, including random intercepts and slopes for all main effects). 
All variables were effect-coded to facilitate interpretation of interactions.
Because the three-way interaction in this model is complex to reason about, we begin by considering a sub-model restricted to the repeated set only.
We found a significant main effect of condition, $b=0.25$,  $t(110.8)=10.2$,  $p < 0.001$, with responses to low prior variability words being more similar at both phases, as well as a main effect of phase $b=-0.38$, $t(120.0)=16.49$, $p <0.001$, with more similar responses in the post-test for all words.
We also find a significant interaction, $b=-0.08$, $t(355.1)=-3.87$, $p=0.0001$, with high prior variability words experiencing an even larger shift, likely reflecting a floor effect.
We evaluate the null hypothesis that this convergence simply reflects additional practice with the task using our control words, which appeared only in the pre-test and post-test. 
We found evidence of a three-way interaction, where the interaction between phase and condition reported above is significantly different from the relationship found for control words, where similarity between partners remained relatively unchanged between the pre- and post-test (Figure~\ref{fig:exp2_results}B). 

\paragraph{The dynamics of adaptation}

\begin{figure}[t!]
\centering
\includegraphics[width=.7\textwidth]{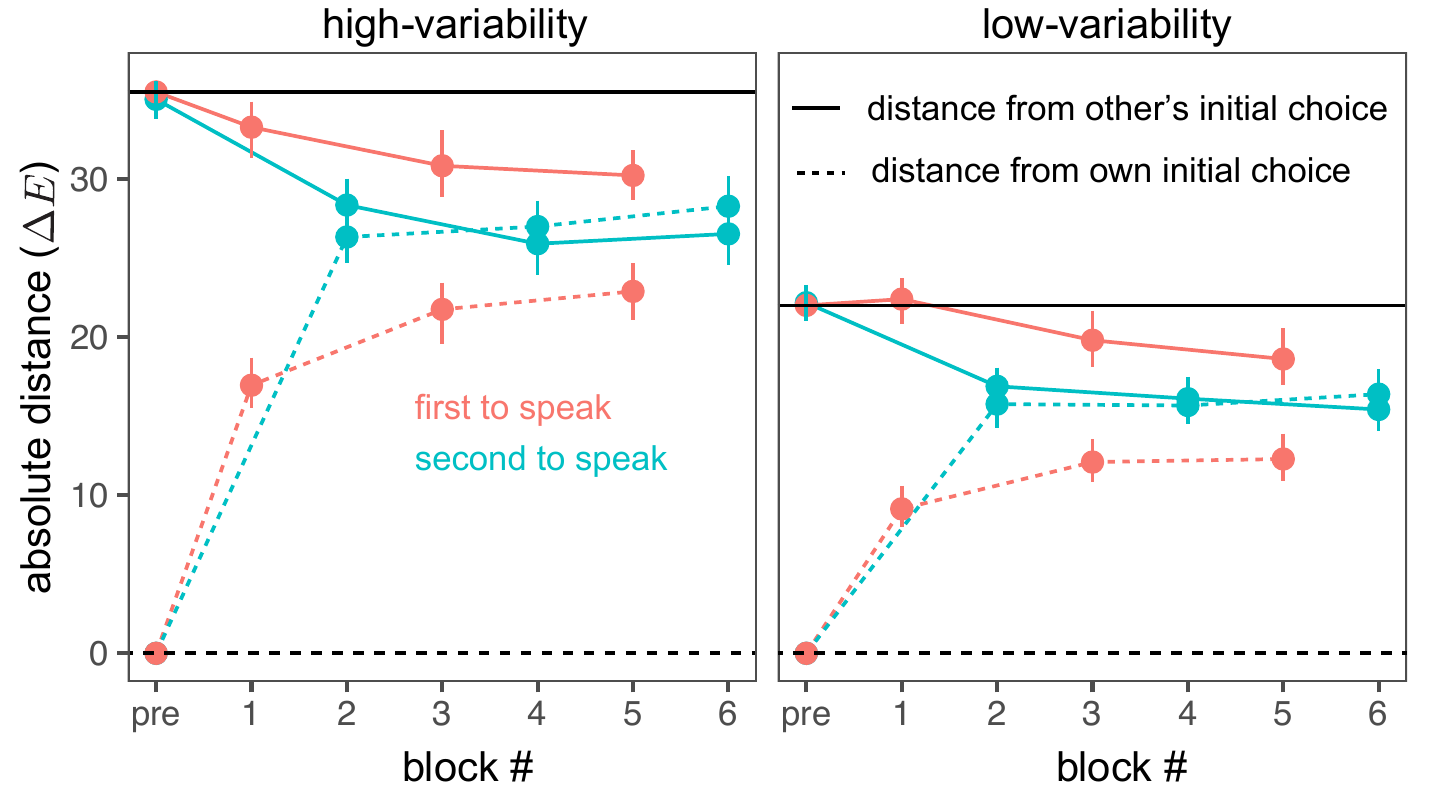}
\caption{\red{\emph{Dynamics of adaption in Exp.~2.} Dotted lines represent distance from one's own initial choice while solid lines represent the distance from one's partner's initial choice choice. The red and blue lines track each participant’s identity as they alternate roles.
Both participants tend to shift away from their initial associations over time (dotted lines rise) while also moving closer to their partner’s initial associations (solid lines fall).
However, there is significant asymmetry in the directionality of adaptation: the first participant to take the role of the speaker shifts from their initial association to a lesser extent than the second participant that takes the role of the speaker. Horizontal black lines are provided for easier comparison to pre-test distances; error bars represent bootstrapped 95\% CIs.}}
\label{fig:exp2_results_directionality}
\end{figure}

While we established strong effects of adaptation from the pre-test to the post-test, it remains unclear how, exactly, adaptation unfolds throughout the communicative interaction.
Our final exploratory analysis examines how each participant's choices as speaker change from round to round as they observe one another's behavior. 
We took advantage of our experimental design, which required participants to alternate between the speaker and listener roles at the outset of each repetition block.
We measured two distances at each block $i$ (see Figure \ref{fig:exp2_results_directionality}): (1) the distance between the chosen color chip chosen by the speaker on that block and their own initial choice as measured in the pre-test (which we call ``distance from own initial'') and (2) the distance between the speaker's chosen color chip and their \emph{partner's} initial choice (which we call ``distance from other's initial''; note that the partner’s initial choice is unobserved). 
We tested the effects of speaker turn (first vs. second), initial variability condition (high vs. low), and block number in mixed-effects regression models predicting these distances, including random intercepts at the subject-level and item-level as well as random effects for speaker turn at the subject level.



First, overall, participants tend to shift away from their initial associations over time, $b=1.2$, $t(4713) = 5.5$, $p<0.001$, while also moving closer to their partner’s initial associations, $b=-1.1,~t(4702) = -4.7$, $p<0.001$.
These shift suggests that partners tend to settle somewhere in the middle of their own initial associations and their partner's initial associations.
Second, however, we find a significant asymmetry in how much each participant adapts. 
The first participant to take the role of the speaker shifts away from their initial association to a lesser extent than the second participant that takes the role of the speaker, $b=-4.9$, $t(111.7) = -6.03$, $p<0.001$ and also shifts closer to their partner to a lesser extent, $b=3.7$, $t(113.8)=5.1$, $p<0.001.$
This asymmetry is consistent with a ``first mover advantage'' where the first participant in the speaker role only has their own \textit{lexical prior} to go on, while the second participant has additional information from observing those initial choices and is able to use that information to anchor their choices closer to their partner's.
The asymmetry created by this alternation is solidified over the remainder of the game: the second participant actually ends up adopting a convention that is closer to their \emph{partner}'s initial association than to their own. 
Third, we find that the initial variability condition appears to simply shift this overall pattern up or down without qualitatively changing the effect.
In the low-variability condition, participants begin closer together but adaptation proceeds similarly, $b=-10.0$, $t(211.4)=-10.8$, $p<0.001$ and $b=-10.3$, $t(216.9)=-9.7$, $p<0.001$ for one's own associations and one's partner's associations, respectively.

\section{Discussion}

Communication is a continual challenge. 
Even when we speak the same language, our vocabularies are full of words that may mean different things to different partners.
In this paper, we investigated the mental representations of meaning that people use to overcome this challenge. 
Using the domain of color-concept associations, where population variability can be carefully measured and manipulated, we first established that variability does not take people by surprise: individuals represent internal uncertainty about associations rather than point estimates and these agreement were well-calibrated to the actual population-wide statistics. 
We then used these elicited distributions to systematically manipulate the degree of variability in the lexical priors of partners in an interactive communication game. 
Although communication was initially difficult for words with more variable associations, participants were able to quickly adapt their expectations based on common ground accumulated within the game, leading their priors to become more similar over time. 
Taken together, these findings suggest that partners enter communicative settings with well-calibrated but flexible priors about the likely difficulty of communication. 
Our work provides new support for recent probabilistic accounts of \emph{lexical uncertainty} in communication and raises a number of new directions. 

First, most prominently, translating our findings from the relatively low-dimensional domain of colors back to the much larger lexical priors for natural language expressions will require further methodological advances.
While the gap between these domains may seem to limit generalization and any potential differences in signaling medium need to be evaluated empirically, we suggest that colors may have more in common with words as a signalling medium than is initially apparent \cite{schloss2018color_inference}. 
Just as the meanings of discrete word ``tokens'' are typically thought to lie in a continuous space (such that we can meaningfully talk about the semantic similarity between words or sentences), participants were presented with a discrete set of 88 color ``tokens'' whose semantic similarity is also grounded in a continuous underlying space (i.e. LAB space). 
Just as natural-language speakers cannot directly convey meanings from the continuous semantic space and must pass through the bottleneck of discrete tokens, our participants could not directly access the continuous color space; they had to pass through the bottleneck of discrete chips.
Perhaps the most significant difference between domains is not discreteness but that the underlying similarity of the chips is exposed visually, whereas the semantic similarity of word tokens is typically not exposed visually (e.g. words that are nearby in meaning often appear very different when written out orthographically). 
Especially because we organized the discrete set of color tokens according to their visual similarity in the task interface, we may have set up an easier pathway for speakers to compare similar signals. 
But we expect that this difference would primarily affect search and retrieval from the space of discrete tokens rather than the representation of uncertainty about a given token’s meaning or the choice to use that token.

Second, our evidence for lexical uncertainty raises deeper mechanistic questions about how, exactly, variability is encoded and learned by individuals.
There are a number of different computational models for probabilistic meanings that have arisen in natural language processing (NLP), including Gaussian embeddings and Gaussian mixtures \cite{DBLP:journals/corr/VilnisM14,athiwaratkun2018probabilistic,bravzinskas2017embedding} as well as a number of different proposals for how these distributions may be learned over time depending on an individual's own idiosyncratic experiences \cite{johns2018large,johns2019using,kraljic2008first,kleinschmidt2019structure}.
A related direction is to better characterize the mechanistic processes allowing participants to align their associations to communicate better (as we found in Experiment 2), and whether updates to an individual's associations are long-lasting or transient. 
One lower-level explanation for alignment is that participants are simply priming one another and making certain associations more salient or accessible  \cite{pickering2006alignment}. 
A higher-level explanation, not mutually exclusive with priming, is that speakers update their beliefs to form partner-specific common ground \cite{Hawkins2021FromPT}.
This hierarchical account predicts that the extent to which local adaptation will persist in longer-term updates depends on sustained use of that association over time, across different partners. 
For example, as a slang term is more consistently and widely used throughout a language community, it may eventually supplant whatever initial associations individuals had with that word and persist as a longer-term update.
However, such persistence would likely require more than a single interaction.

A third direction for future work is to examine how sources of variability in color associations arise in the first place. 
While we confirmed a coarse relationship between the abstractness of a concept and the variability of its associations, it remains unclear how this relationship arises.
On one hand, some portion of this relationship may be driven by sensory aspects of word representations. 
We are all exposed to roughly the same visual imagery statistics for concrete concepts such as ``tree,'' suggesting strong associations with greens and browns.
Meanwhile, abstract concepts such as ``justice'' lack concrete referents and associations may be driven by more idiosyncratic semantic properties that vary depending on each individuals' own history with the concept.
These terms may be more susceptible to cultural variation across different languages and latent social groups \cite{Tham2019ASI}, which would be interesting to measure in broader cross-cultural samples.

It is worth noting several potential limitations of our data.
For one, there are intrinsic challenges associated with accurately sampling responses from color space.
While we evenly sampled color chips from the Munsell color chart \cite{munsell1907color,Landa2005ChartingCF}, following standard practice for color elicitation \cite{brown1954study,berlin1991basic,gibson2017color,sturges1995locating}, there are known distortions created by this set of chips \cite{zaslavsky2018efficient}.
Most noticeably, discrepancies in the relative number of green-blue and red-pink chips compared to yellow-ish chips may have biased responses towards better-represented hues.
We expected this distortion to have the biggest effect on \emph{entropy-based} measures of variability, which is more sensitive to the support of the response distribution then our $\Delta E$ measure.
Second, our online data collection setup prevented us from ensuring perfectly consistent color calibration across participants' screens, raising concerns that variability in associations is simply due to presentation noise. 
While we did our best to minimize such sources of variability and bias, our primary comparisons were importantly \emph{relative} comparisons between different words (e.g. between words with more or less variability). 
Because the display of colors on a given participant's screen was likely to be fixed across the experiment, any noise or bias arising from the color display should contribute equally across all words and conditions.
Third, it is possible that our use of within-participant designs, both when eliciting multiple responses across blocks in Experiment 1 and when interleaving high and low variability contexts in Experiment 2, may have resulted in spillover or ``self-priming'' effects that reduce our estimates of internal variability.
Generally, these possible effects would work against our hypotheses: in Experiment 1, self-priming would have favored the point estimate hypothesis rather than the internal uncertainty hypothesis, and in Experiment 2, it would have reduced our estimate of differences across conditions. 
More broadly, it will be important to reproduce our findings using longer delays between blocks.

More broadly, color associations are of substantial interest for communication in their own right. 
The very properties that we highlighted in Experiment 1 may be responsible for the prevalence and usefulness of color in communicating about abstractions, relative to more concrete modalities \cite{gass2014being,johansson2020color,winter2019sensory,schloss2020blue}.
When a speaker says ``love is blue'' they draw attention to different semantic dimensions than ``love is bright red,'' which may be difficult to reach with other metaphorical expressions.
Thus, characterizing uncertainty in communication with color associations is not only useful for practical \emph{visual communication} like the choice of color scales in data visualization \cite{lin2013selecting,setlur2015linguistic,schloss2020semantic}, or for design elements like signage \cite{mahnke1996color,schloss2018color}, but is core to a more general theory of multi-modal communication.
The meanings of colors may have much in common with the meanings of words we use in everyday conversations.
Rather than rigid dictionaries, well-calibrated probability distributions allow us to better anticipate misunderstandings and coordinate with one another to achieve mutual understanding.

\section{Acknowledgements}

We are grateful for helpful conversations with Bill Thompson, Christiane Fellbaum, Josh Peterson, and Ken Norman. This work was supported by NSF Grant \#1911835 to RH. 

\section{Author Contributions}

All authors conceived of the project and contributed to the study design. SM and RH implemented the experiments, collected data, performed all analyses, and wrote the manuscript. TG provided feedback on the manuscript. All authors approved the final version of the manuscript for submission.

\bibliographystyle{apacite}

\setlength{\bibleftmargin}{.125in}
\setlength{\bibindent}{-\bibleftmargin}

\bibliography{refs}

\renewcommand{\thefigure}{S\arabic{figure}}
\renewcommand{\thetable}{S\arabic{table}}
\setcounter{table}{0}
\setcounter{figure}{0}

\newpage
\section*{Appendix A: What is a lexical prior?}

While the present work focuses on basic  \emph{qualitative} properties of lexical priors, and a formal model is not strictly necessary to interpret our findings, an overview of previous formal definitions nonetheless helps make this construct more precise \cite<see>[for a more extensive treatment]{bergen2016pragmatic, Hawkins2021FromPT}.
We begin by defining a \emph{lexicon} $\mathcal{L}$, which specifies a specific set of form-meaning mappings, assign semantic meanings to all  utterances in a language. 
\begin{definition}
A \emph{lexicon} $\mathcal{L} : (u, m) \rightarrow \{0,1\}$ is a function assigning a Boolean truth value\footnote{This definition can be generalized to a real-valued function $\mathcal{L} : (u, m) \rightarrow \mathbb{R}$ \cite{degen2020redundancy}, yielding a graded or ``fuzzy'' semantics, but we use the traditional truth-conditional for simplicity.} to every pair of utterances $u\in \mathcal{U}$ and meanings $m\in \mathcal{M}$.
\end{definition}
Traditionally, all speakers of a language are assumed to learn a single fixed lexicon specifying the meaning of every utterance in the language. 
\begin{example}
Consider a simple referential language game where there are 2 possible utterances $u \in \{u_1,u_2\}$ and 2 possible meanings $m \in \{m_1, m_2\}$. 
Then a single lexicon $\mathcal{L}$ can be represented as a binary $|\mathcal{U}| \times |\mathcal{M}|$ matrix with 2 rows and 2 columns, where each entry represents whether utterance $u$ has meaning $m$ or not, and $\mathcal{L}(u,w)$ simply looks up the specified entry:
$$
\mathcal{L} = \begin{pmatrix}
   1  & 0   \\
   1 & 1 
\end{pmatrix}
$$
\end{example}

Now, it is straightforward to define a lexical prior as a probability distribution over such matrices, representing an agent's uncertainty over exactly which lexicon is being used by their partner.
\begin{definition}
A \emph{lexical prior} is a probability distribution over the support of possible lexicons, denoted by $P(\mathcal{L})$.
\end{definition}
Note that the lexical prior $P(\mathcal{L}$ is distinct from $P(u)$ or $P(m)$, which simply represent the background probability of a given utterance or meaning independently popping up in the environment. 

\begin{example}
Let $\mathcal{L}_{ij}$ be the matrix entry representing whether utterance $u_i$ has meaning $m_j$.
Then $$P(\mathcal{L}_{ij}) \sim \textrm{Bernoulli}(.5)$$ defines a maximally uninformative prior where every utterance is equally likely \emph{a priori} to have every meaning. 
\end{example}
\begin{example}
If we additionally assume that lexicons satisfy the constraint that entries sum to one for each column of the matrix (a weak form of mutual exclusivity, where every meaning is assumed to have exactly utterance that expresses it), then the lexical prior may be specified even more compactly in conditional form as $$P(\mathcal{L}_{ij}) = P(u_i | m_j) \sim Categorical(\theta_j),$$ 
where every meaning $m_j$ is associated with a vector $\theta_j$ giving a distribution over utterances.
\end{example}

We are now prepared to consider how these theoretical constructs correspond our setting of color-concept signaling games, where the meanings $m$ are a discrete set of concept words (e.g. \emph{lemon}, \emph{randomness}) and the utterances $u$ are a discrete set of 88 color chips. 
We could in principle try to elicit the full latent object $\mathcal{L}$ by showing participants every color-concept pair $(u, m)$ and collecting a slider response or 2AFC judgement about whether that participant (or that participant’s partner) would endorse that specific mapping. 
This kind of elicitation procedure might be the most direct instantiation of the formal definition but is expensive to collect (e.g. requiring $50 \times 88 = 4400$ judgements from a single participant to cover just a quarter of the concepts we use). 
Instead, we elicit the prior by taking samples from the conditional distribution $P(u | m)$, which  also happens to be a well-vetted method for eliciting color associations.
In other words, we query a column of the lexicon (i.e. conditioning on a given concept) and ask participants to draw a sample from the induced distribution over color chips.
For this reason, it is convenient to define the lexical prior \emph{for a given concept $m$} as the elicited distribution over colors given $m$, although it is a slight abuse of the term.



\section*{Appendix B: Supplemental figures \& code}

\begin{figure}[H]
\centering
\includegraphics[width=\textwidth]{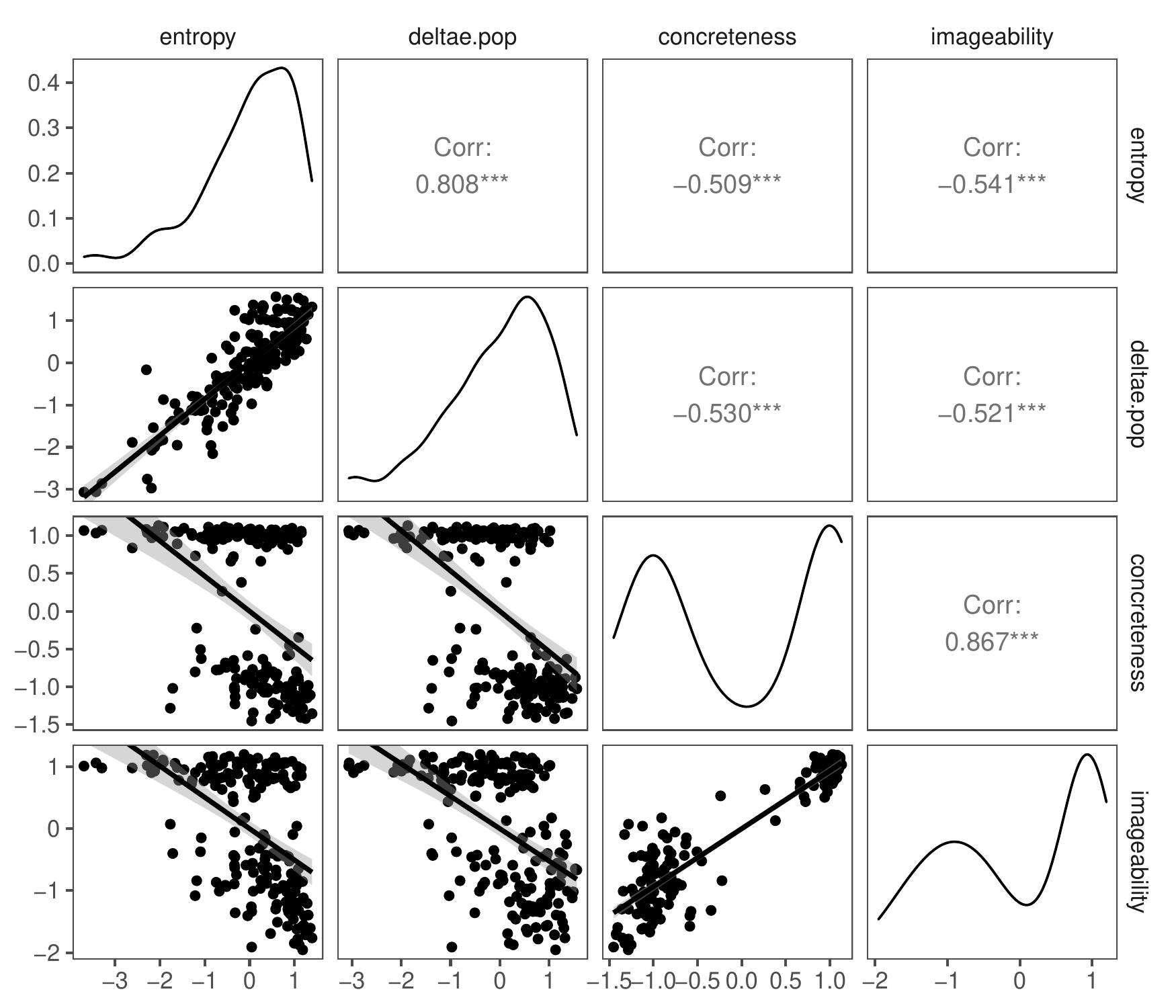}
\caption{\emph{Pairwise relationships between variables in concreteness and imageability analyses.} Values in upper triangular represent Spearman rank correlations. Measures of variability are highly correlated with one another ($\rho=0.81$), as are the semantic measures of concreteness and imageability ($\rho=0.87$).}
\label{fig:relationships}
\end{figure}

\newpage

\begin{figure}
\begin{lstlisting}
 Family: bernoulli 
  Links: mu = logit 
Formula: correct ~ condition * blockNum + (1 + condition * blockNum | gameId) + (1 + blockNum | target) 
   Data: . (Number of observations: 5136) 
Samples: 4 chains, each with iter = 2000; warmup = 1000; thin = 1;
         total post-warmup samples = 4000

Group-Level Effects: 
~gameId (Number of levels: 107) 
                                                  Estimate Est.Error l-95% CI u-95% CI Rhat Bulk_ESS
sd(Intercept)                                         0.62      0.12     0.39     0.86 1.00     2141
sd(conditionconcrete)                                 1.24      0.20     0.86     1.65 1.00     1045
sd(blockNum)                                          0.29      0.05     0.20     0.38 1.00     1754
sd(conditionconcrete:blockNum)                        0.25      0.10     0.04     0.44 1.01      744
cor(Intercept,conditionconcrete)                     -0.36      0.19    -0.68     0.06 1.00      820
cor(Intercept,blockNum)                               0.56      0.21     0.12     0.91 1.00      721
cor(conditionconcrete,blockNum)                      -0.13      0.19    -0.49     0.24 1.00     1494
cor(Intercept,conditionconcrete:blockNum)            -0.21      0.30    -0.76     0.38 1.00     1619
cor(conditionconcrete,conditionconcrete:blockNum)     0.34      0.30    -0.28     0.86 1.00     1564
cor(blockNum,conditionconcrete:blockNum)             -0.05      0.30    -0.61     0.57 1.00     1763
                                                  Tail_ESS
sd(Intercept)                                         2831
sd(conditionconcrete)                                 2170
sd(blockNum)                                          2748
sd(conditionconcrete:blockNum)                         954
cor(Intercept,conditionconcrete)                      1531
cor(Intercept,blockNum)                               1524
cor(conditionconcrete,blockNum)                       2898
cor(Intercept,conditionconcrete:blockNum)             2542
cor(conditionconcrete,conditionconcrete:blockNum)     1929
cor(blockNum,conditionconcrete:blockNum)              2682

~target (Number of levels: 199) 
                        Estimate Est.Error l-95% CI u-95% CI Rhat Bulk_ESS Tail_ESS
sd(Intercept)               0.58      0.10     0.40     0.77 1.00     2226     2765
sd(blockNum)                0.12      0.05     0.02     0.21 1.01      460      714
cor(Intercept,blockNum)     0.48      0.34    -0.23     0.97 1.01      873     1570

Population-Level Effects: 
                           Estimate Est.Error l-95% CI u-95% CI Rhat Bulk_ESS Tail_ESS
Intercept                      0.13      0.12    -0.10     0.36 1.00     4240     3377
conditionconcrete              0.87      0.20     0.48     1.27 1.00     3154     3309
blockNum                       0.42      0.05     0.33     0.52 1.00     3024     2347
conditionconcrete:blockNum     0.22      0.08     0.07     0.38 1.00     2718     2762

Samples were drawn using sampling(NUTS). For each parameter, Bulk_ESS
and Tail_ESS are effective sample size measures, and Rhat is the potential
scale reduction factor on split chains (at convergence, Rhat = 1).
\end{lstlisting}
\caption{Bayesian regression output for Experiment 2, using the maximal random effect structure.}
\label{lst:bayesianmodel}
\end{figure}


\begin{figure}
\centering
\begin{subfigure}{\textwidth}
  \centering
  \includegraphics[width=\linewidth]{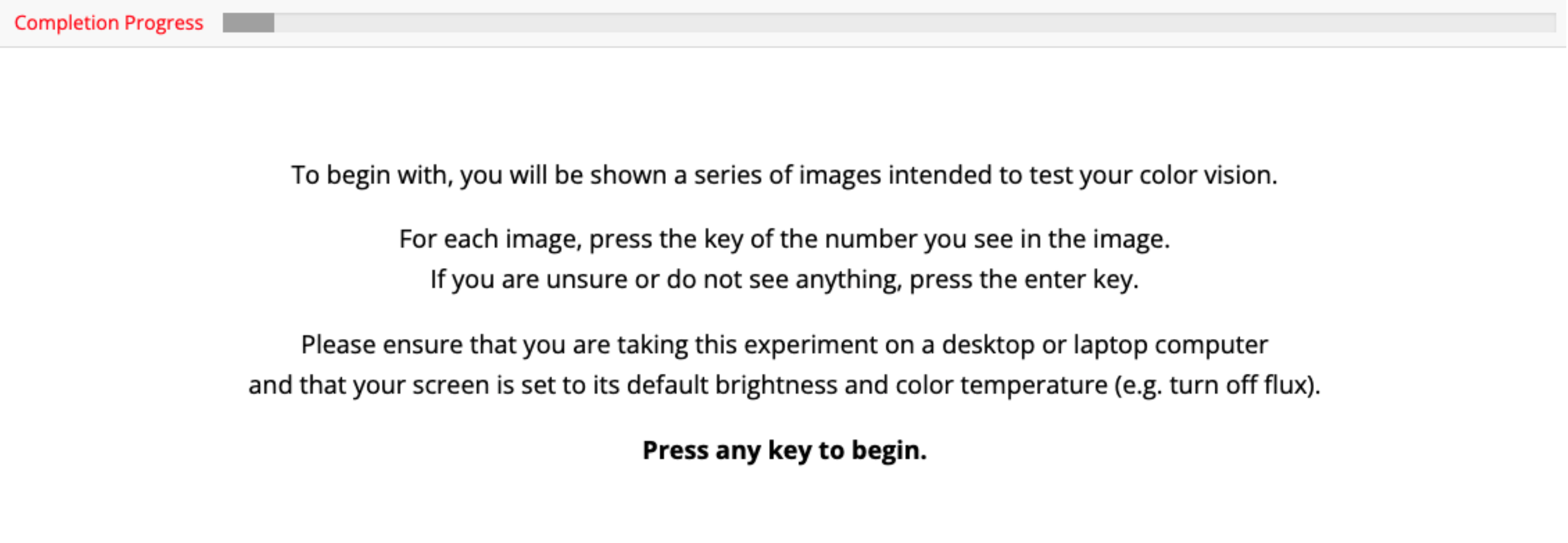}
\end{subfigure}%
\newline
\begin{subfigure}{.5\textwidth}
  \centering
  \includegraphics[width=\linewidth]{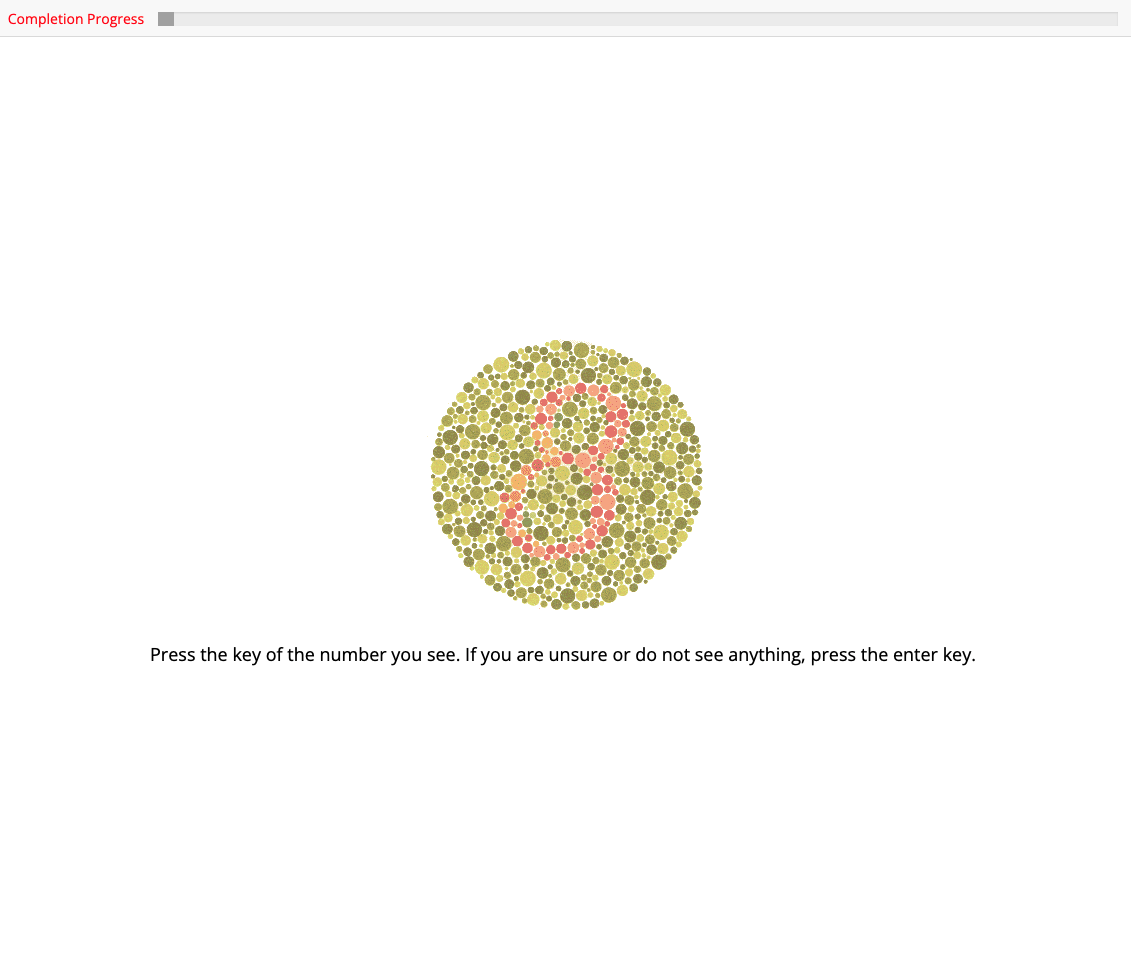}
\end{subfigure}%
\begin{subfigure}{.5\textwidth}
  \centering
  \includegraphics[width=\linewidth]{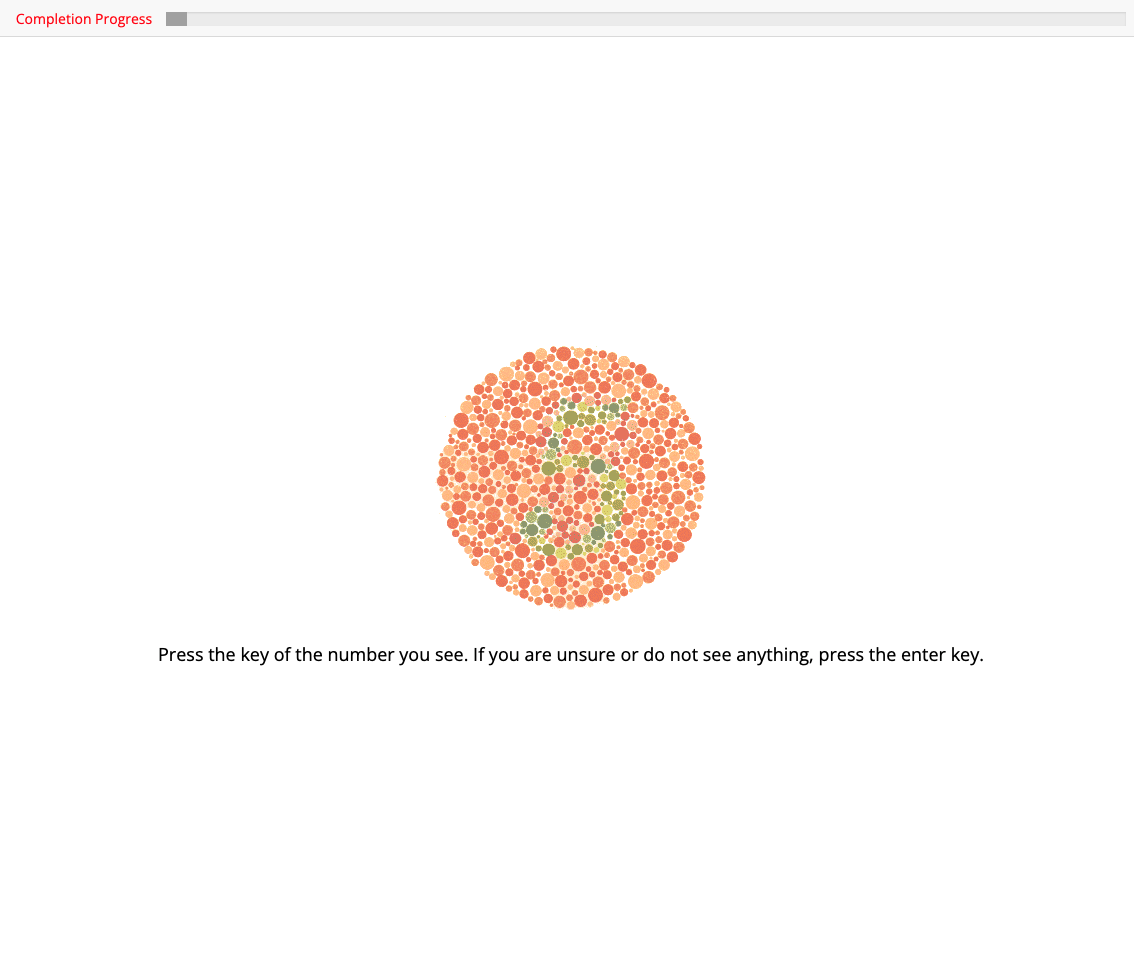}
\end{subfigure}
\caption{\emph{Sample color vision trials for Experiments 1}. Participants were presented with three Ishihara plates selected to differentiate those with normal color vision from those with red-green deficiencies or more extreme colorblindness. Participants were instructed to enter the number they saw, or nothing, if they were unsure or did not see anything. The same plates were presented to participants in Experiment 2 in the form of a pre-test comprehension quiz.}
\label{fig:colorblind-trials}
\end{figure}

\begin{figure}
\centering
\begin{subfigure}{\textwidth}
  \centering
  \includegraphics[width=\linewidth]{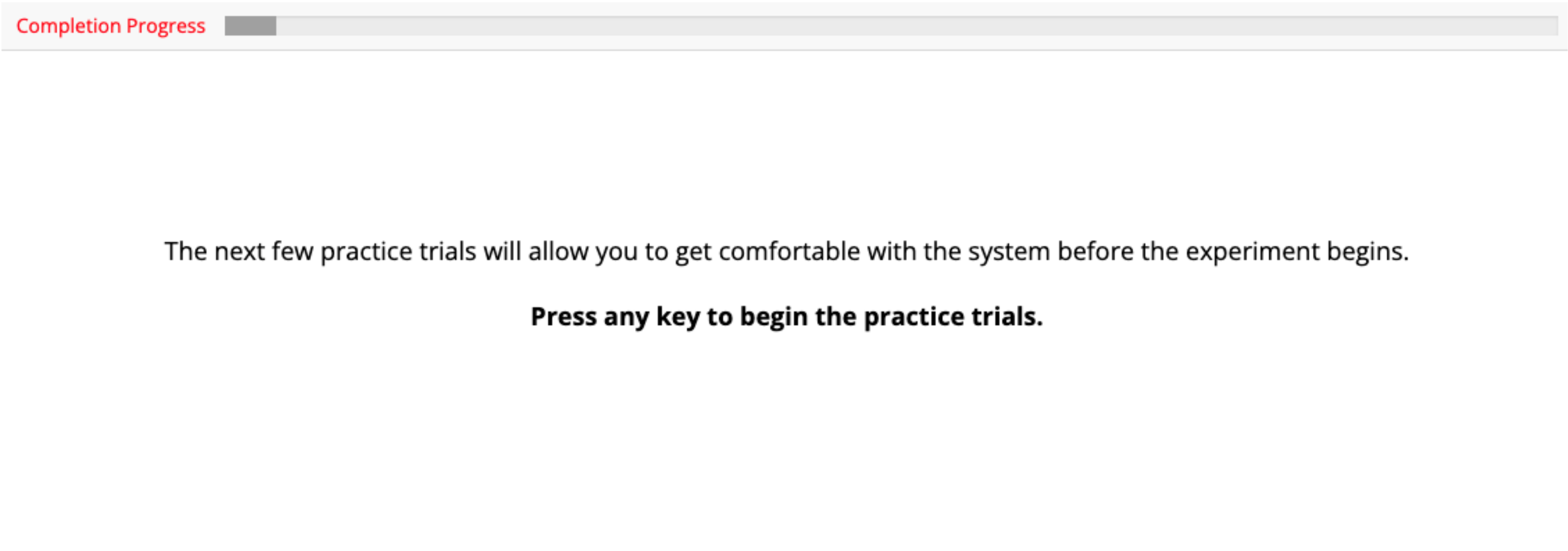}
\end{subfigure}%
\newline
\begin{subfigure}{.5\textwidth}
  \centering
  \includegraphics[width=\linewidth]{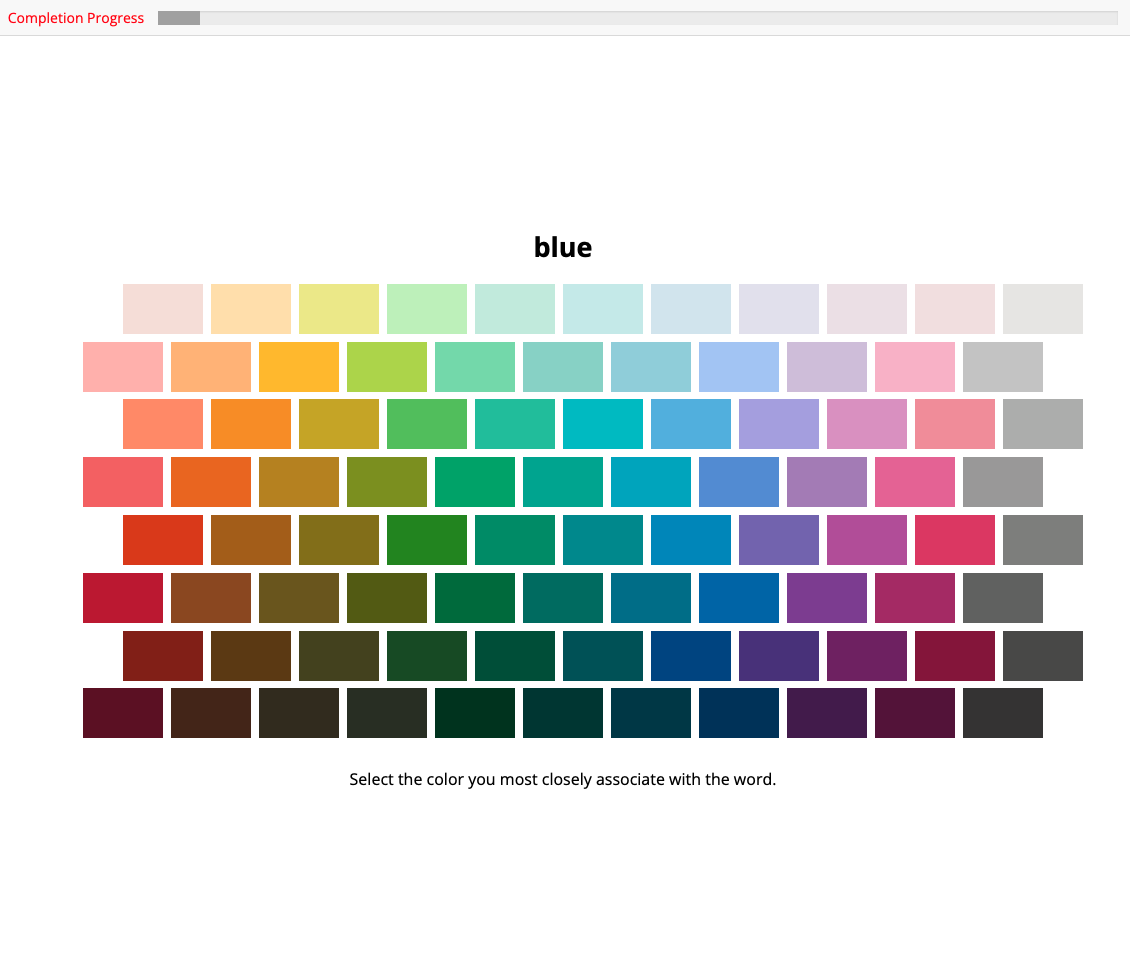}
\end{subfigure}%
\begin{subfigure}{.5\textwidth}
  \centering
  \includegraphics[width=\linewidth]{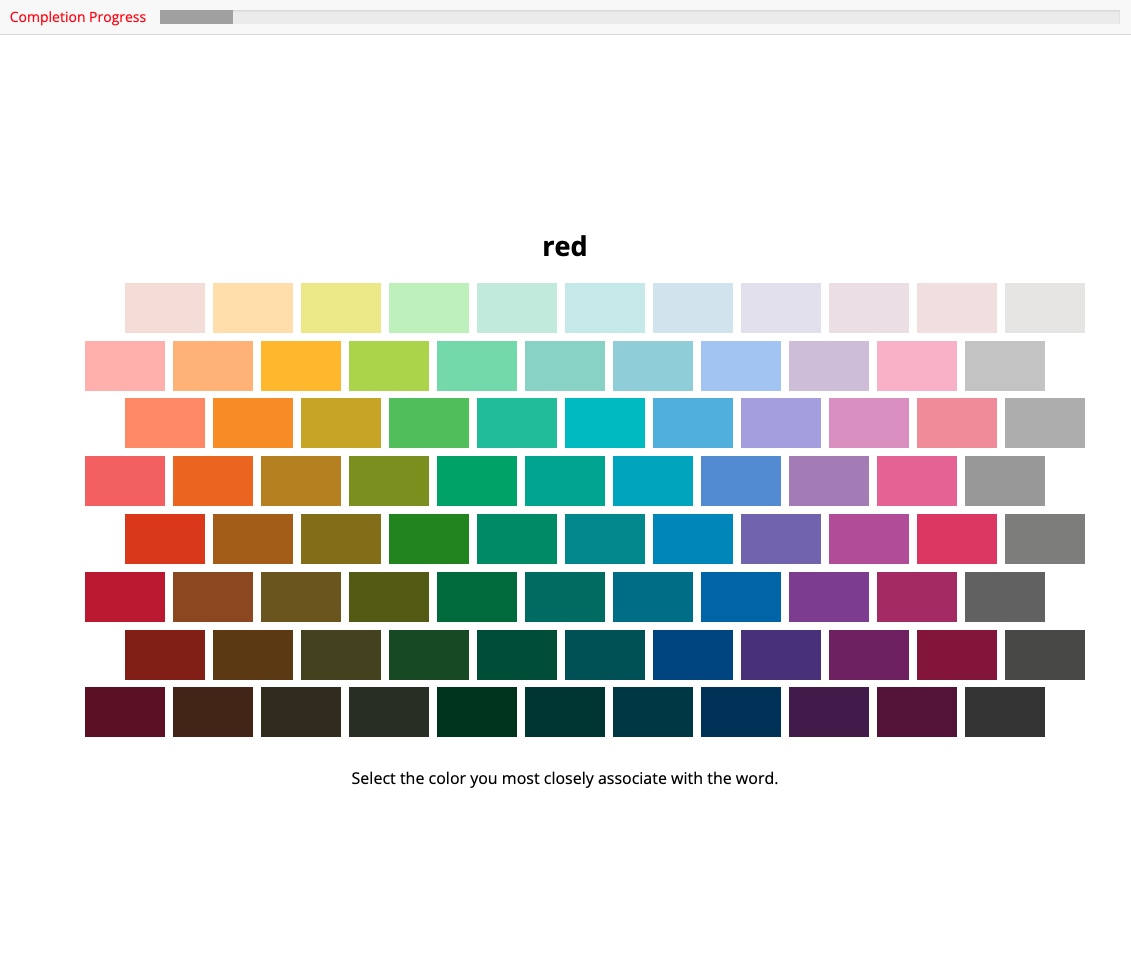}
\end{subfigure}
\caption{\emph{Sample pre-test trials for Experiment 1}. Participants were asked to select color swatches for color words like ‘blue’ and ‘red’ to ensure that any differences in color displays lay within tolerance of color boundaries.}
\label{fig:colorword-trials}
\end{figure}

\begin{figure}[h]
\centering
\includegraphics[width=0.9\textwidth]{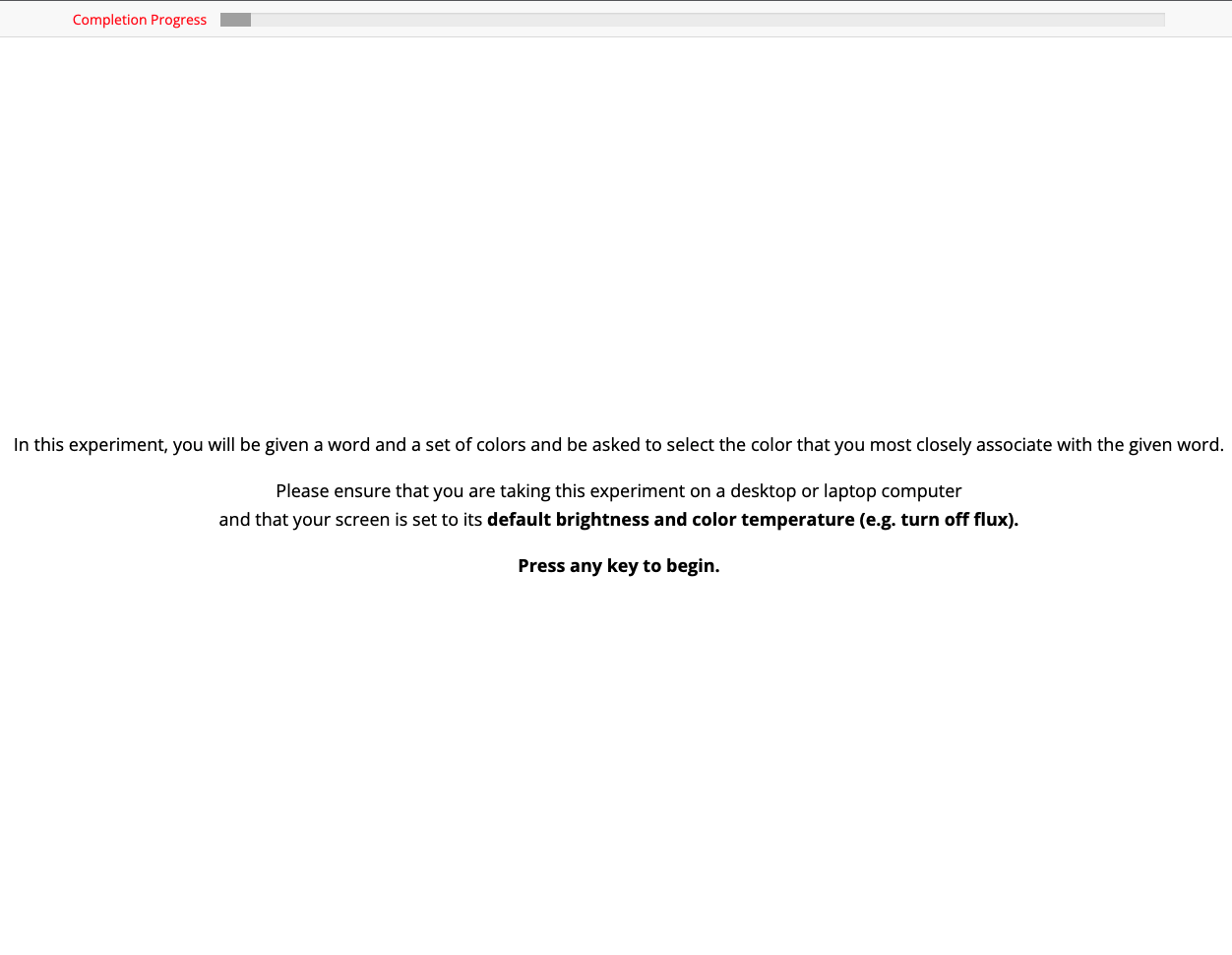}
\caption{\emph{Instructions for main trials of Experiment 1}. To mitigate variance in individual’s color displays, participants were instructed to use desktop or laptop computers rather and ensure that their screens were set to the default brightness and color temperature (e.g. to turn off Flux). }
\label{fig:instructions-experiment1}
\end{figure}

\begin{figure}
\centering
\includegraphics[width=.8\textwidth]{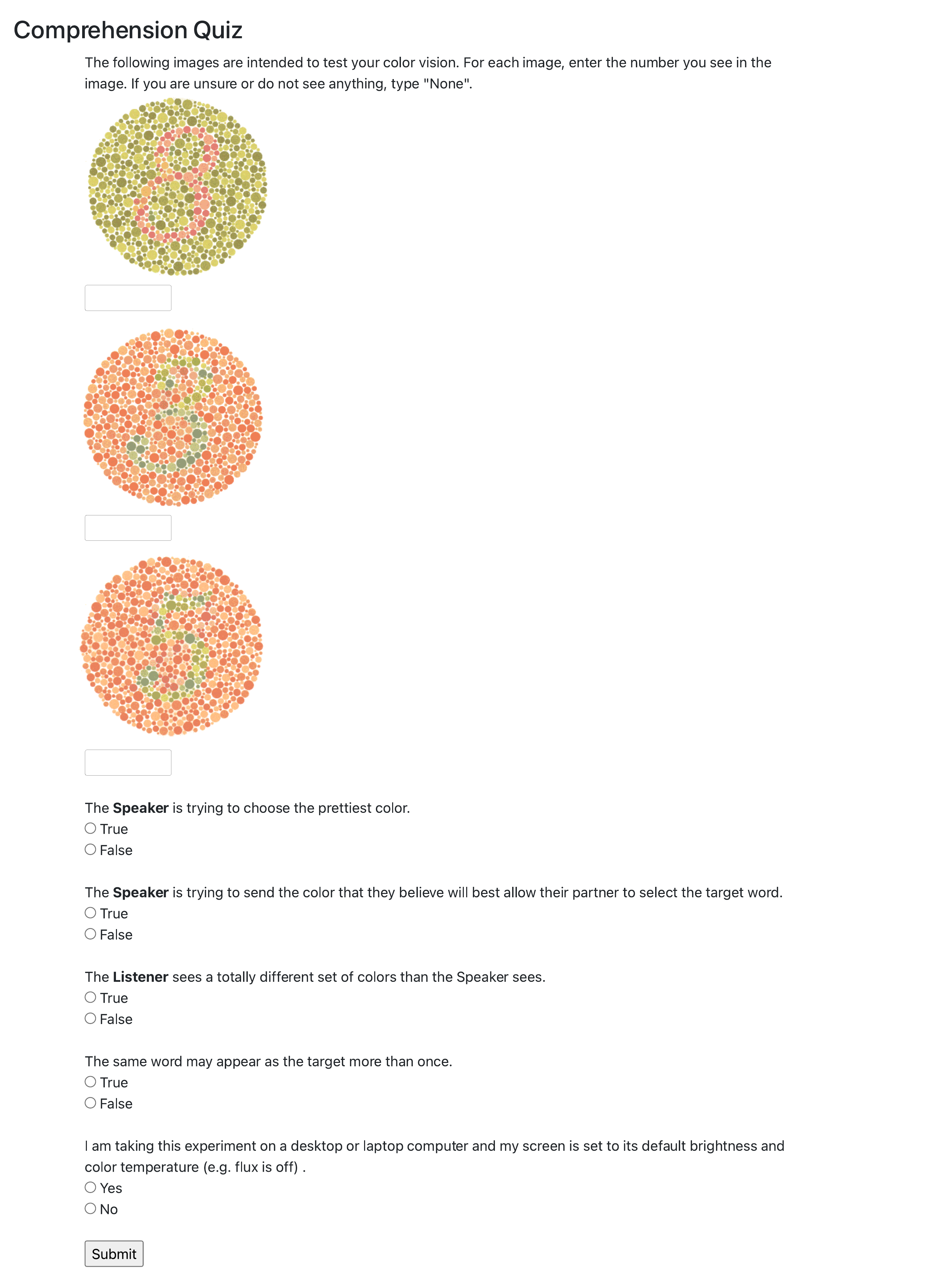}
\caption{\emph{Pre-test comprehension quiz for Experiment 2}. Participants demonstrated their understanding of the study instructions and desired display settings through a pre-test comprehension quiz. Color vision was also tested using Ishihara plates as in Experiment 1.}
\label{fig:exp2-comprehension}
\end{figure}

\end{document}